\documentclass{article} 
\usepackage{iclr2026_conference,times}
\usepackage{subcaption}

\usepackage{amsmath, amssymb, amsthm, bbm}

\def\eqref#1{equation~\ref{#1}}

\def\1{\bm{1}}

\DeclareMathAlphabet{\mathsfit}{\encodingdefault}{\sfdefault}{m}{sl}
\SetMathAlphabet{\mathsfit}{bold}{\encodingdefault}{\sfdefault}{bx}{n}

\usepackage{amsthm}
\usepackage{pifont}
\usepackage{csquotes}
\usepackage[T1]{fontenc}

\usepackage[table,dvipsnames]{xcolor}
\usepackage{subcaption}
\newcolumntype{H}{>{\setbox0=\hbox\bgroup}c<{\egroup}@{}} 
\usepackage{hyperref}
\usepackage{url}

\title{Policy Optimization Prefers The Path of Least Resistance}

\iclrfinalcopy

\author{Debdeep Sanyal\thanks{Work done while at RespAI, KIIT Bhubaneswar.}\\
Birla AI Labs\\
Bhubaneswar, India\\
\texttt{debdeep.sanyal-st@adityabirla.com}\\
\And
Aakash Sen Sharma\\
InvideoAI\\
Mumbai, India\\
\texttt{aakash.sensharma@invideo.io}\\
\And
Dhruv Kumar \\
BITS Pilani \\
Rajasthan, India\\
\texttt{dhruv.kumar@pilani.bits-pilani.ac.in}\\
\And
Saurabh Deshpande \\
Birla AI Labs \\
Germany \\
\texttt{saurabh.deshpande-c@adityabirla.com}\\
\And
Murari Mandal \\
Kalinga Institute of Industrial Technology\\
Bhubaneswar, India\\
\texttt{murari.mandalfcs@kiit.ac.in}
}

\usepackage{graphicx}
\usepackage{amsmath}
\usepackage{amssymb}
\usepackage{booktabs}
\usepackage{multirow}
\usepackage{multicol}
\usepackage{array}
\usepackage{float}
\usepackage{mdframed}
\usepackage{algorithm}
\usepackage{algpseudocode}
\usepackage[skip=1pt]{caption}
\usepackage{lipsum}
\usepackage[most]{tcolorbox}

\begin{document}
\maketitle
\newcommand{\myparatight}[1]{\smallskip\noindent{\bf {#1}.}~}
\tcbset{
    userstyle/.style={
        enhanced,
        colback=white,
        colframe=black,
        colbacktitle=gray!20,
        coltitle=black,
        rounded corners,
        boxrule=0.5pt,
        drop shadow=black!50!white,
        attach boxed title to top left={
            xshift=-2mm,
            yshift=-2mm
        },
        boxed title style={
            rounded corners,
            size=small,
            colback=gray!20
        },
        fontupper=\footnotesize,
        left=1mm,
        right=1mm,
        top=2mm,
        bottom=1mm
    },
    jailbreakstyle/.style={
        enhanced,
        colback=white,
        colframe=red,
        colbacktitle=red!40,
        coltitle=black,
        rounded corners,
        sharp corners=north,
        boxrule=0.5pt,
        drop shadow=red!50!white,
        attach boxed title to top left={
            xshift=-2mm,
            yshift=-2mm
        },
        boxed title style={
            rounded corners,
            size=small,
            colback=red!20
        },
        fontupper=\footnotesize,
        left=1mm,
        right=1mm,
        top=2mm,
        bottom=1mm
    },
    jailbreakstyleres/.style={
        enhanced,
        colback=white,
        colframe=red,
        colbacktitle=red!40,
        coltitle=black,
        rounded corners,
        sharp corners=north,
        boxrule=0.5pt,
        drop shadow=red!50!white,
        attach boxed title to top right={
            xshift=-2mm,
            yshift=-2mm
        },
        boxed title style={
            rounded corners,
            size=small,
            colback=red!0
        },
        fontupper=\footnotesize,
        left=1mm,
        right=1mm,
        top=2mm,
        bottom=1mm
    },
    myreplyborderstyle/.style={
        enhanced,
        colback=white,
        colframe=black,
        colbacktitle=red!40,
        coltitle=black,
        rounded corners,
        sharp corners=north,
        boxrule=0.5pt,
        drop shadow=black!50!white,
        attach boxed title to top right={
            xshift=-2mm,
            yshift=-2mm
        },
        boxed title style={
            rounded corners,
            size=small,
            colback=red!0
        },
        fontupper=\footnotesize,
        left=1mm,
        right=1mm,
        top=2mm,
        bottom=1mm
    },
    replystyleg/.style={
        enhanced,
        colback=blue!0,
        colbacktitle=black,
        colframe=black,
        coltitle=black,
        boxrule=1pt,
        drop shadow=black!50!,
        rounded corners,
        sharp corners=north,
        attach boxed title to top right={
            xshift=-2mm,
            yshift=-2mm
        },
        boxed title style={
            rounded corners,
            size=small,
            colback=blue!0,
        },
        fontupper=\footnotesize,
        left=1mm,
        right=1mm,
        top=2mm,
        bottom=1mm
    },
    replystyler/.style={
        enhanced,
        colback=blue!15,
        colframe=black,
        colbacktitle=blue!20,
        coltitle=black,
        boxrule=0.5pt,
        drop shadow=black!50!white,
        rounded corners,
        sharp corners=north,
        attach boxed title to top right={
            xshift=-2mm,
            yshift=-2mm
        },
        boxed title style={
            rounded corners,
            size=small,
        },
        fontupper=\footnotesize,
        left=1mm,
        right=1mm,
        top=2mm,
        bottom=1mm
    },
    replystylew/.style={
        enhanced,
        colback=purple!5,
        colframe=black,
        colbacktitle=pink!40,
        coltitle=black,
        boxrule=0.5pt,
        drop shadow=black!50!white,
        rounded corners,
        sharp corners=north,
        attach boxed title to top right={
            xshift=-2mm,
            yshift=-2mm
        },
        boxed title style={
            rounded corners,
            size=small,
            colback=pink!60
        },
        fontupper=\footnotesize,
        left=1mm,
        right=1mm,
        top=2mm,
        bottom=1mm
    }
}

\newtcolorbox{userquery}[1][]{
    userstyle,
    title=Prompt,
    #1
}

\newtcolorbox{llmreply-g}[1][]{
    replystyleg,
    title=Response,
    #1
}

\newtcolorbox{llmreply-r}[1][]{
    replystyler,
    title=Response,
    #1
}

\newtcolorbox{mybox}[2][]{
    replystyler,
    title=#2,
    #1
}
\newtcolorbox{myboxw}[2][]{
    replystylew,
    title=#2,
    #1
}

\newtcolorbox{myboxg}[2][]{
    replystyleg,
    title=#2,
    #1
}

\newtcolorbox{myuser}[2][]{
    userstyle,
    title=#2,
    #1
}

\newtcolorbox{myjailbreak}[2][]{
    jailbreakstyle,
    title=#2,
    #1
}

\newtcolorbox{myreplyborder}[2][]{
    myreplyborderstyle,
    title=#2,
    #1
}

\renewcommand{\paragraph}[1]{\noindent\textbf{#1~}}
\begin{abstract}
Policy optimization (PO) algorithms are used to refine Large Language Models (LLMs) for complex, multi-step reasoning. Current state-of-the-art pipelines enforce a strict think-then-answer format to elicit chain-of-thought (CoT); however, the behavior of PO when these rigid constraints are relaxed into an open-ended CoT structure remains an under-studied question. We investigate this gap with an extensive suite of controlled experiments and identify a consistent principle: \textit{policy optimization consistently follows the path of least resistance}. When afforded the flexibility to interleave reasoning and response, policy optimization consistently learns to discard explicit reasoning, causing the policy to degenerate to a direct \texttt{<answer>}-only format. This outcome holds true across a rigorous evaluation suite spanning 5 model families (4B-12B), 3 reasoning domains (math, code, logic), and 3 distinct PO algorithms (GRPO, DAPO, REINFORCE++). We find that this collapse in format is persistent even when the complex \texttt{<think><answer>} format is assigned up to 4x larger reward weights. We formalize this principle through a series of controlled reward decomposition experiments, demonstrating a clear hierarchy: PO systematically optimizes for the simplest reward component first, a preference that holds even when faced with mutually exclusive choices or strong incentives for more complex behaviors. Finally, we show that successful convergence on the high-reward shortcut is not a low-effort drift but is driven by the optimization process that requires the KL-regularized policy to have sufficient freedom to make a significant shift from its initial prior. Our findings reveal that granting policies the freedom to diverge is a double-edged sword: while necessary for discovering high-reward shortcuts, it also creates a powerful incentive to game the simplest aspects of the reward function, posing a critical challenge for reward hacking under alignment.
\end{abstract}
\section{Introduction}
Policy Optimization \cite{dapo, grpo, drgrpo, vapo, prorl} has emerged as the principal tool for refining Large Language Models \cite{gemma, qwen, mistral, llama} towards complex, multi-step reasoning. The community's dominant approach is one of careful enforcement: to elicit a chain-of-thought, we engineer a rigid reward function that strongly compels the model to follow a strict \enquote{think-then-answer} format \cite{prorl, deepseek, apple, longpo}. While this enforced structure is effective in producing a desired output, it obscures a deeper, more fundamental question that has been largely overlooked: What is the optimizer's \textit{innate preference} when the external guidance is removed? If we grant the model the freedom to choose its own path to a solution, what path does it take? This question is not merely academic. If the optimizer possesses a powerful intrinsic bias, then our current methods of alignment may be working against a fundamental force, leading to brittle, inefficient, and unpredictable training dynamics. Understanding this preference is therefore a critical, yet under-studied, prerequisite for building truly robust and reliable reasoning systems.

Our investigation, therefore, began with a simple act of liberation. We designed a composite reward function that, for the first time, offered the model a genuine choice. Instead of a single, valid path, our function rewarded any number of interleaved \texttt{<think>} and \texttt{<answer>} blocks, and crucially, also rewarded a direct \texttt{<answer>}-only format. This seemingly minor change from enforcement to choice led to a striking anomaly. Across a rigorous evaluation suite spanning diverse architectures (Gemma-3 \cite{gemma}, Qwen-2.5 \cite{qwen}, Llama-3.1 \cite{llama}, Ministral \cite{mistral}, Yi \cite{yi}), scales (4B to 12B), and PO algorithms (GRPO \cite{grpo}, DAPO \cite{dapo}, REINFORCE++ \cite{reinforce}), the policy invariably discarded complexity. On challenging domains from mathematics (GSM8K, MathHard) to coding (rStar-Coder) and logic (ReClor), the outcome was identical: the structured thoughts vanished, and the model converged on the simplest possible path. This powerful, emergent preference for a \enquote{Cognitive Shortcut} became the central mystery we sought to solve.

\begin{figure}[t]
    \centering
    \includegraphics[width=\linewidth]{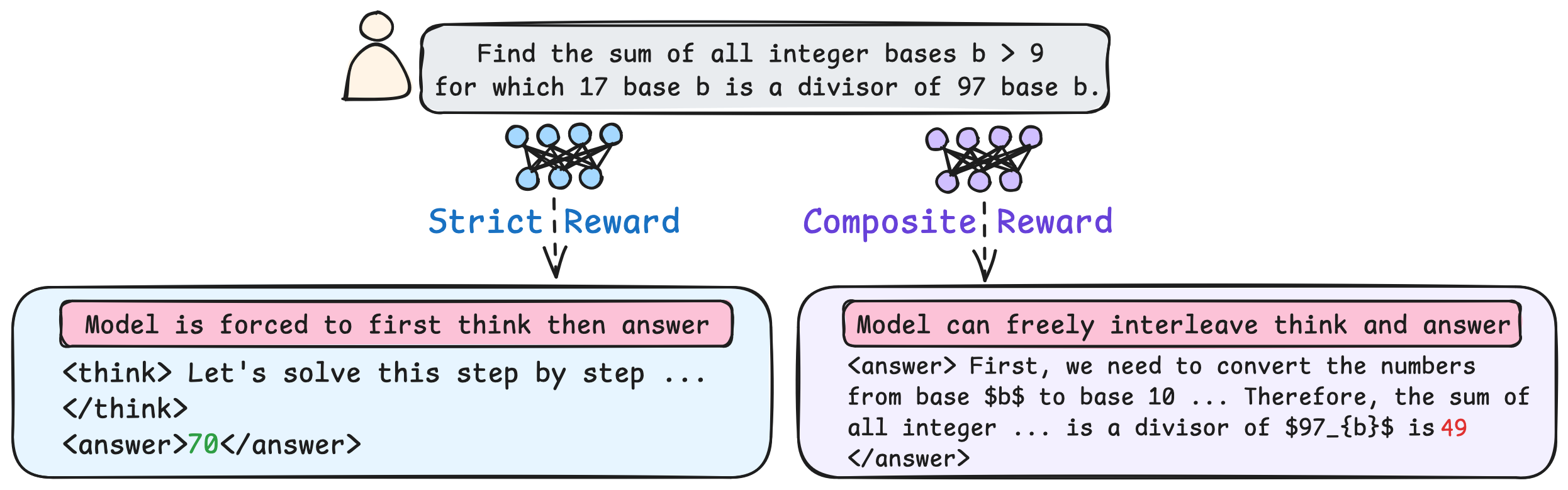}
    \caption{
         We compare the standard Strict Reward paradigm \textbf{(left)}, which enforces a rigid \textit{think-then-answer} structure, against our novel Composite Reward \textbf{(right)}, which grants the model the freedom to choose its solution path. The central finding investigated in this paper is the emergent policy learned under this freedom: the model consistently bypasses explicit reasoning to converge on the simplest valid format, a phenomenon we term the \textit{Cognitive Shortcut.}}
    \label{fig:LTT}
    \vspace{-1\baselineskip}
\end{figure}

This initial finding sparked a natural line of inquiry. The model's preference for the simplest format was clear, but was this an \enquote{all-or-nothing} choice, or was there a more nuanced structure to this preference? Figure~\ref{fig:LTT} This led us to our next hypothesis: if the optimizer is biased towards simplicity, perhaps it tackles complex objectives not holistically, but by first conquering their simplest components. To test this, we moved from a simple choice to a structured hierarchy. We designed a controlled \enquote{Reward Hierarchy} experiment using three nested reward formats of ascending difficulty $\textcolor{blue}{r_1} < \textcolor{ForestGreen}{r_2} < \textcolor{orange}{r_3}$, all yielding the same reward magnitude. The result was a stunningly predictable sequence of learning: the optimizer first mastered the simplest format $\textcolor{blue}{r_1}$, and only after this reward was saturated did it begin to make progress on $\textcolor{ForestGreen}{r_2}$. The most complex format, $\textcolor{orange}{r_3}$, was never learned. This revealed that the \enquote{Principle of Least Resistance} is not just a preference, but a sequential law.\par

This discovery, however, raised an even more pressing question: just how powerful is this law? Is it a mere tie-breaker, or a dominant force that can override other incentives? To quantify its strength, we returned to our reward hierarchy, but this time we offered exponentially larger rewards for mastering the more complex formats. We found that the model would consistently forgo significant rewards to remain on the simpler path. Only when the incentive for complexity became overwhelmingly large did we observe a \textit{phase transition} where the optimizer was finally bribed into tackling the harder task. The preference for simplicity, we realized, was a powerful, quantifiable force in the optimization landscape.

Finally, with the behavioral law and its strength firmly established, we turned to the ultimate question: why does this law exist? This question has taken on a new urgency. A prominent trend in state-of-the-art policy optimization is the removal of conservative constraints like the KL penalty \cite{dapo, vapo}, granting models unprecedented freedom to explore the reward landscape. The common intuition is that this freedom simply allows for more effective reward maximization \cite{prorl}. Our investigation, however, reveals a more complex and cautionary reality. By treating the KL divergence as a diagnostic for this exploratory freedom, we found that successful convergence on the \textit{Cognitive Shortcut} is not a low-effort drift. Instead, it is an aggressive optimization process that requires a large and decisive policy shift away from the model's pre-trained priors. The path of least resistance is not the path of lowest policy shift, but the path carved by the most powerful and stable gradient signal, a form of reward hacking. This finding suggests that unconstrained exploration, while powerful, may have unintended and problematic consequences.

Our work makes the following contributions:

\ding{182} \textbf{A New Fundamental Principle:} We identify, formalize, and empirically validate the \textit{Principle of Least Resistance}, a powerful predictive law governing the behavior of policy optimization in LLMs.

\ding{183} \textbf{Rigorous Ablation Testing:} We move beyond simple observation, stress-testing our principle with a motivated sequence of novel, controlled experiments (Reward Hierarchies, Exponential Gambits) that quantify its strength and universality.

\ding{184} \textbf{A Counter-Intuitive KL Perspective:} We provide a new  lens for understanding PO dynamics, demonstrating that successful convergence on simple shortcuts requires a high-KL policy shift, thereby linking learnability to the freedom to escape the inertia of priors.

\section{Setup}
To rigorously test our hypothesis, we designed a comprehensive experimental suite. Our methodology was guided by two core principles: \textbf{diversity}, to ensure our findings are general and not artifacts of a specific domain or model; and \textbf{relevance}, to use tasks that are widely recognized as benchmarks for complex, multi-step reasoning. We selected six powerful, publicly available models from five distinct architectural families, with scales ranging from 4 billion to 24 billion parameters. Our suite included Gemma-3 (4B \& 12B) \cite{gemma}, Qwen-2.5 (7B) \cite{qwen}, Llama-3.1 (8B) \cite{llama}, Ministral (8B) \cite{mistral}, and Yi (6B) \cite{yi}. This diversity ensures our conclusions are not an artifact of a specific model's pre-training or architecture but are a general property of these systems. All experiments were conducted on a cluster of 3 NVIDIA RTX A6000 GPUs, with 48GB VRAM each, and a single NVIDIA H100 GPU, with 80GB VRAM.

\textbf{Datasets.} To rigorously test our \enquote{Principle of Least Resistance}, we curated a diverse gauntlet of datasets spanning three critical reasoning domains. For mathematical reasoning, we used a tiered selection from the foundational multi-step arithmetic of \textbf{GSM8K} \cite{gsm8k} to the more complex algebraic challenges in \textbf{Math-Hard} \cite{mathhard} and the rich, SOTA traces of \textbf{Open-R1 Math 220k} \cite{openr1}. To evaluate algorithmic and code reasoning, we leveraged the competition-level complexity of \textbf{Microsoft's rStar-Coder} \cite{msoft} alongside the functionally verifiable problems in \textbf{Open R1 verifiable coding} \cite{openr1}. Finally, for logical and deductive reasoning, we tested our models on the canonical suppositional puzzles of \textbf{Knights and Knaves} \cite{knk}, the natural language deduction required by \textbf{ReClor} \cite{reclor}, and the sequential problem-solving of a specialized \textbf{planning-mystery} dataset. This multi-domain, multi-difficulty suite was designed to ensure that our findings are a general principle of optimization, not an artifact of a single task.
\textbf{Policy Optimization Algorithms.} Our findings are not an artifact of a single training algorithm. To establish that the \enquote{Principle of Least Resistance} is a feature of the PO paradigm itself, not a quirk of one implementation, we replicated our experiments across three distinct and powerful families of policy optimization algorithms. Our selection was designed to cover a range of modern techniques, from highly-engineered systems to bias-corrected standards.

First, we employed \textbf{DAPO (Decoupled clip and Dynamic sampling Policy Optimization)} \cite{dapo}, a state-of-the-art system designed for stable, large-scale training of reasoning models. Its objective function is given by:
\begin{equation}
\begin{split}
\mathcal{J}_{\text{DAPO}}(\theta) &= \mathbb{E}_{(q,a) \sim \mathcal{D}, \{o_i\}_{i=1}^G \sim \pi_{\theta_{\text{old}}}(\cdot|q)}\\ & \left[ \frac{1}{\sum_{i=1}^G |o_i|} \sum_{i=1}^G \sum_{t=1}^{|o_i|} \right.\quad \left. \min\left(r_{i,t}(\theta)\hat{A}_{i,t}, \text{clip}(r_{i,t}(\theta), 1-\epsilon_{\text{low}}, 1+\epsilon_{\text{high}})\hat{A}_{i,t}\right) \right]
\end{split}
\end{equation}
\begin{equation}
\text{s.t. } 0 < |\{o_i \mid \text{is\_equivalent}(a, o_i)\}| < G
\end{equation}

We further used \textbf{Dr. GRPO (Group Relative Policy Optimization Done Right)} \cite{drgrpo}. We specifically chose the \enquote{Dr.} variant over the original GRPO for its response length and question difficulty bias correction, which provides a more stable and accurate learning signal. Its objective is:
$$
\mathcal{J}_{\text{Dr.GRPO}} = \frac{1}{G} \sum_{i=1}^{G} \frac{1}{|o_{i}|} \sum_{t=1}^{|o_{i}|} \min \left\{ r_{i,t}(\theta) \hat{A}_{i,t}, \text{clip} \left( r_{i,t}(\theta), 1-\epsilon, 1+\epsilon \right) \hat{A}_{i,t} \right\}
$$
where the advantage $\hat{A}_{i,t}$ is a per-sequence reward baseline: $\hat{A}_{i,t} = R(\mathbf{q}, \mathbf{o}_{i}) - \text{mean}(\{R(\mathbf{q}, \mathbf{o}_{1}), \dots, R(\mathbf{q}, \mathbf{o}_{G})\})$.

Finally, to connect our findings to the foundational principles of policy gradients, we included \textbf{REINFORCE++} \cite{reinforce}, a robust and modern variant of the classic REINFORCE algorithm. The consistent emergence of our observed phenomenon across these three distinct algorithmic philosophies provides strong evidence that the \enquote{Path of Least Resistance} is a fundamental property of the policy optimization paradigm, independent of the specific implementation.
\section{Path of Least Resistance}
\subsection{Flexible Format Reward}
\label{sec:flex_format_4.1}
\begin{table}
\centering
\caption{
    \textbf{Performance Comparison of Training Paradigms Across Diverse Models and Reasoning Tasks.}
    We report the final accuracy (\%) of models trained with a \textbf{Strict Reward} (enforcing the \texttt{<think><answer>} format) versus our \textbf{Composite Reward} (allowing a choice of formats). The results are notably mixed across all four reasoning domains, with no single paradigm consistently outperforming the other. This ambiguity suggests that a simple accuracy comparison is insufficient to understand the underlying learning dynamics, motivating a deeper investigation into the optimizer's intrinsic preferences. Best performance in each pair is highlighted in bold.\\
}
\label{tab:t1}
\resizebox{\textwidth}{!}{%
\begin{tabular}{lcccccccc}
\toprule
& \multicolumn{2}{c}{\textbf{GSM8K}} & \multicolumn{2}{c}{\textbf{rStar-Coder}} & \multicolumn{2}{c}{\textbf{ReClor}} & \multicolumn{2}{c}{\textbf{Planning-Mystery}} \\
\cmidrule(lr){2-3} \cmidrule(lr){4-5} \cmidrule(lr){6-7} \cmidrule(lr){8-9}
\textbf{Model} & \textbf{Strict} & \textbf{Composite} & \textbf{Strict} & \textbf{Composite} & \textbf{Strict} & \textbf{Composite} & \textbf{Strict} & \textbf{Composite} \\
\midrule
Gemma-3 4B & 72.4 & \textbf{73.1} & \textbf{55.8} & 53.5 & \textbf{73.2} & 69.4 & \textbf{58.3} & 55.9 \\
Qwen-2.5 7B & \textbf{92.4} & 85.5 & 73.2 & \textbf{73.6} & \textbf{80.1} & 73.8 & \textbf{77.5} & 73.8 \\
Llama-3.1 8B & \textbf{86.2} & 82.9 & \textbf{64.1} & 59.2 & \textbf{78.3} & 72.7 & \textbf{74.9} & 70.6 \\
Ministral 8B & \textbf{89.5} & 84.8 & \textbf{74.0} & 72.6 & \textbf{81.0} & 76.5 & 76.8 & \textbf{77.3} \\
Yi 6B & \textbf{85.1} & 83.8 & \textbf{68.3} & 63.7 & \textbf{68.4} & 65.1 & \textbf{71.2} & 67.7 \\
Gemma-3 12B & \textbf{94.6} & 86.4 & \textbf{76.1} & 72.3 & \textbf{85.5} & 81.2 & \textbf{81.2} & 76.5 \\
\bottomrule
\end{tabular}%
}
\end{table}
Our investigation begins with a simple yet profound departure from the status quo. The prevailing methodology in policy optimization for reasoning tasks enforces a rigid structure on the model, effectively mandating a specific computational path. We hypothesized that this enforcement might be obscuring the optimizer's intrinsic biases. To test this, we asked a fundamental question: \textit{What path does the optimizer choose when it is given a choice?}

To answer this question, we first needed to formally define the choice. Let a model generation be a sequence of tokens $y$. The standard, \textbf{Strict} reward function, $R_{\text{strict}}(y)$, provides a positive reward only if $y$ perfectly matches the \texttt{think-then-answer} format:
\begin{equation}
R_{\text{strict}}(y) =
\begin{cases}
1 & \text{if } y \in \texttt{ <think>.*</think>\textbackslash s*<answer>.*\textbackslash boxed\{.*\}</answer>\$} \\
0 & \text{otherwise}
\end{cases}
\end{equation}

This function defines a single, narrow path to success. To create a choice, we designed a \textbf{Composite} reward function, $R_{\text{composite}}(y)$, which defines a much larger set of valid, rewarded formats. This function allows for any number of interleaved \texttt{<think>} and \texttt{<answer>} blocks, and crucially, also accepts a direct \texttt{<answer>}-only format without any preceding thought:

\begin{equation}
R_{\text{regex}}(y) =
\begin{cases}
1 & \text{if } y \in
    \begin{aligned}[t]
        & \texttt{\char`^(((<think>.*</think>\textbackslash s*<answer>.*</answer>\textbackslash s*)+|} \\
        & \texttt{(<answer>.*</answer>\textbackslash s*(<think>.*</think>} \\& \texttt{\textbackslash s*<answer>.*\textbackslash boxed\{.*\}</answer>\textbackslash s*)*))\$}
    \end{aligned} \\
0 & \text{otherwise}
\end{cases}
\end{equation}
Our initial expectation was that the model, now liberated from its strict format, would learn a nuanced policy, perhaps using longer \texttt{<think>} blocks for harder problems and skipping them for easier problems. The reality was far more dramatic and revealing.

\subsection{Universal Convergence to Minimal-Format Solutions}
Our setup in Section~\ref{sec:flex_format_4.1} was designed to test a core hypothesis: that a model, freed from the constraint of a single solution format, would learn a nuanced, adaptive reasoning policy. The results, presented in Table \ref{tab:t1}, decisively falsify this hypothesis and reveal an alternative organizing principle.

A primary observation from Table \ref{tab:t1} is the consistent and significant performance degradation when models are trained with the $R_{\text{composite}}$ on tasks that demonstrably benefit from structured reasoning. On \textbf{GSM8K}, a benchmark for multi-step arithmetic, the $R_{\text{strict}}$ policy outperforms the $R_{\text{composite}}$ policy by a substantial margin across all models, with performance gaps as large as \textbf{7.1\%} for Qwen-2.5 7B (92.4 vs. 85.5) and \textbf{8.2\%} for Gemma-3 12B (94.6 vs. 86.4). Similar trends are observed on \textbf{rStar-Coder} and \textbf{ReClor}, where the enforced \texttt{think-then-answer} structure provides a clear advantage.

This performance gap is not an indictment of the model's capability, but rather a direct consequence of the policy it learns. When trained with the $R_{\text{composite}}$, the optimizer does not learn a dynamic balance of thinking and answering. Instead, in all experimental runs, \textit{the policy invariably converges to the simplest valid format: the direct \texttt{<answer>}-only response}. This learned behavior, which we term the \textbf{Cognitive Shortcut}, involves the complete omission of the rewarded, and often necessary, intermediate reasoning steps. The model, in optimizing $R_{\text{composite}}$, discovers that the path of least resistance is to forgo the complex, high-utility \texttt{think} block, even when doing so is detrimental to final performance.

We find that the optimizer does not act as a neutral maximizer of expected reward across all valid solution formats. It is a highly biased agent that, when presented with a choice, will aggressively converge on the simplest possible specification of a rewarded behavior. This discovery of the Cognitive Shortcut is not the conclusion of our work, but the foundational anomaly that motivates a deeper, more controlled investigation into the nature and strength of this preference.

\section{Formalizing the Principle of least resistance}

\begin{figure*}[t]
  \centering
  \setlength{\tabcolsep}{2pt} 
  \renewcommand{\arraystretch}{0} 
  \begin{tabular}{ccc}
    \includegraphics[width=0.32\textwidth]{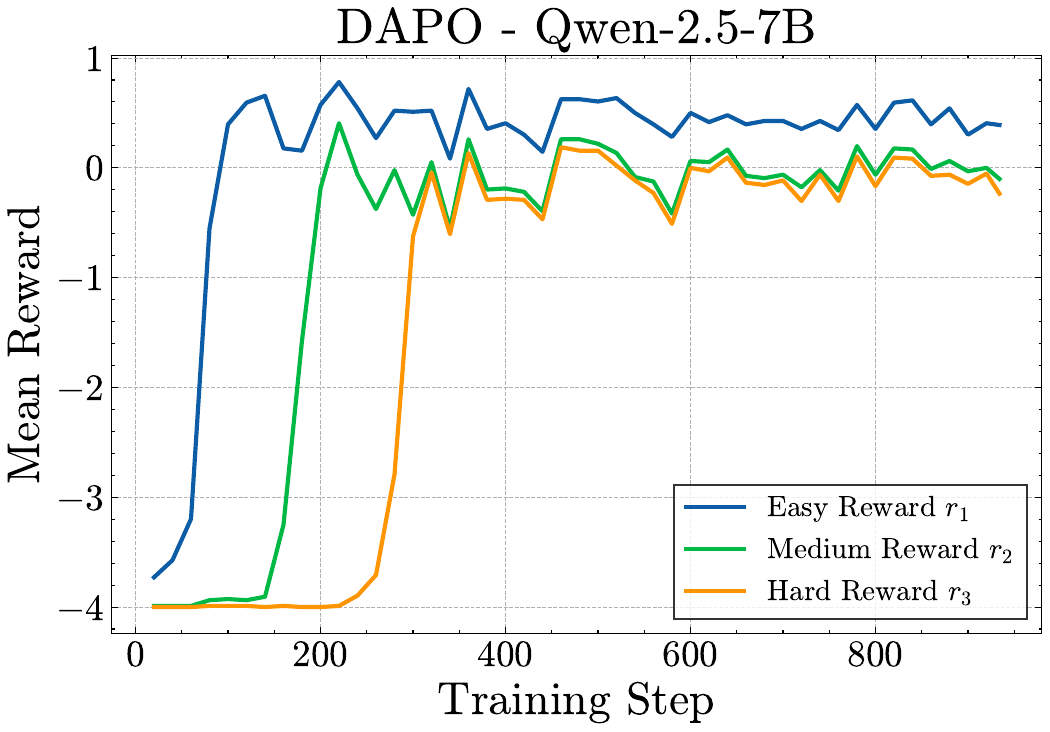} &
    \includegraphics[width=0.32\textwidth]{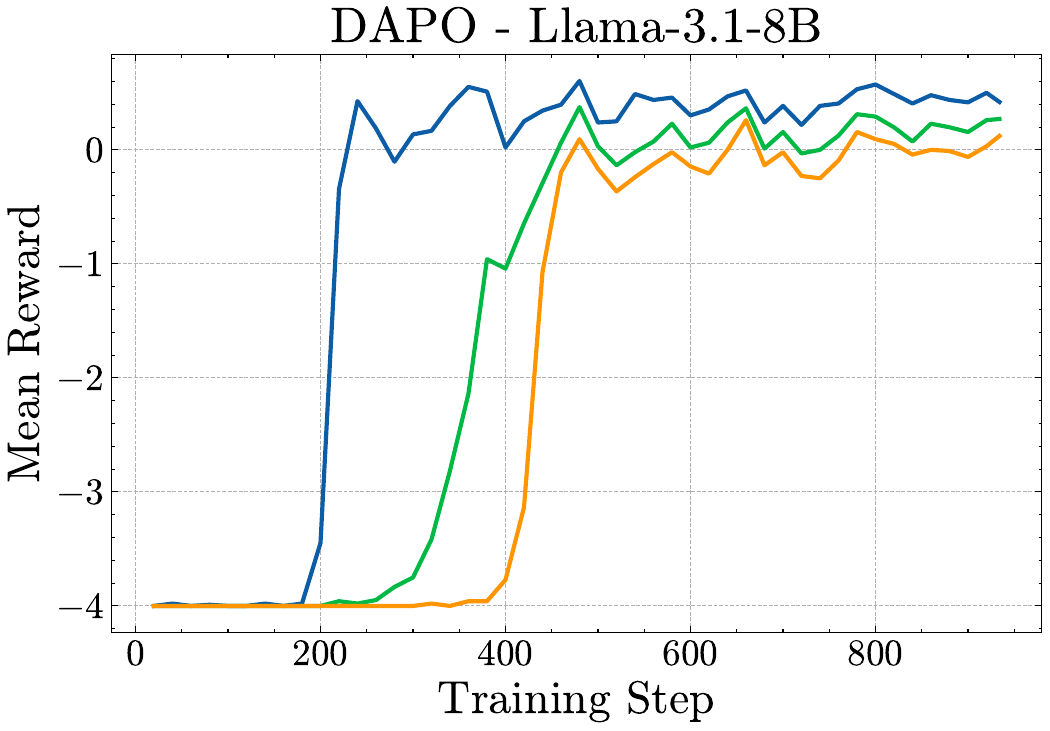} &
    \includegraphics[width=0.32\textwidth]{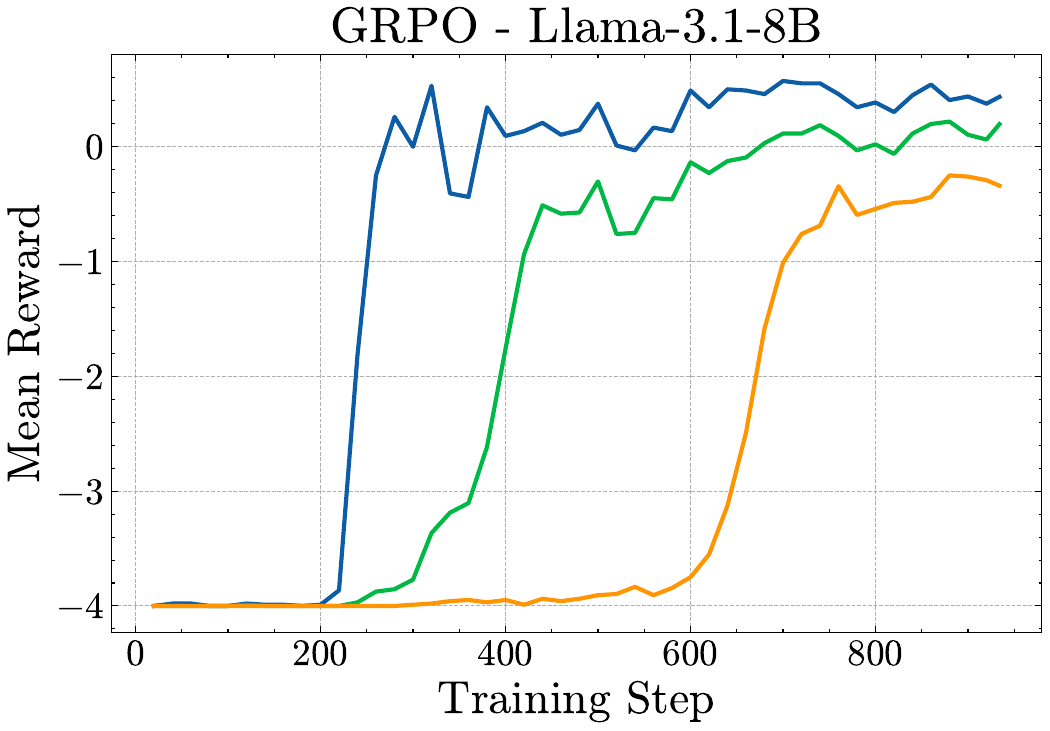} \\
    \includegraphics[width=0.32\textwidth]{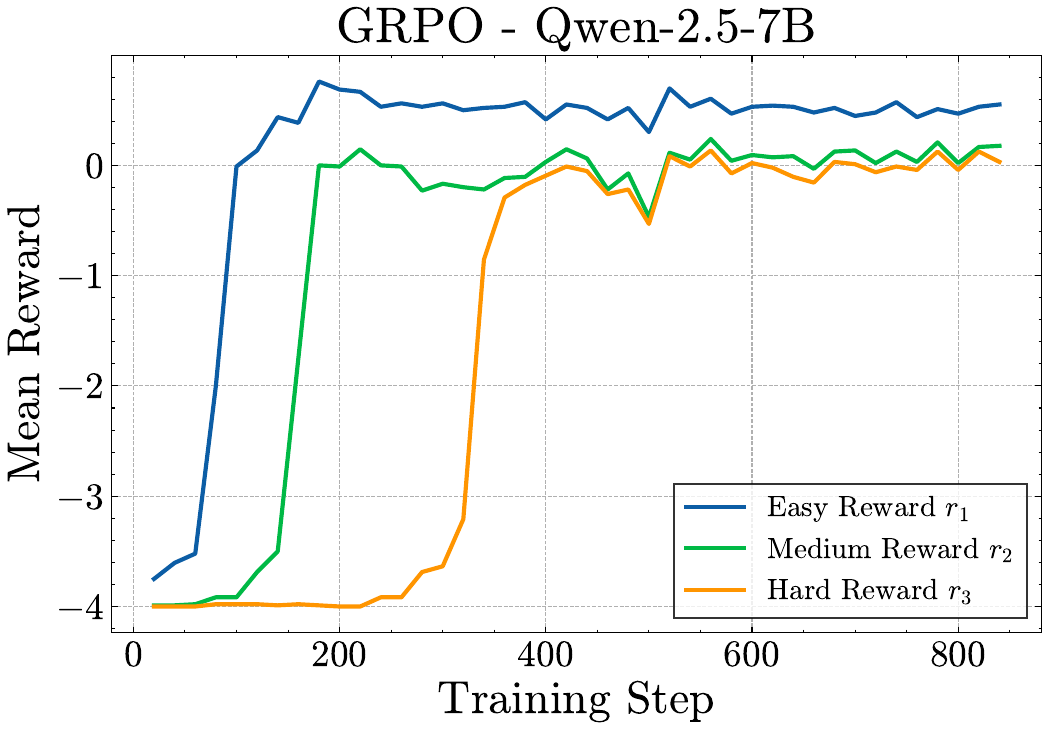} &
    \includegraphics[width=0.32\textwidth]{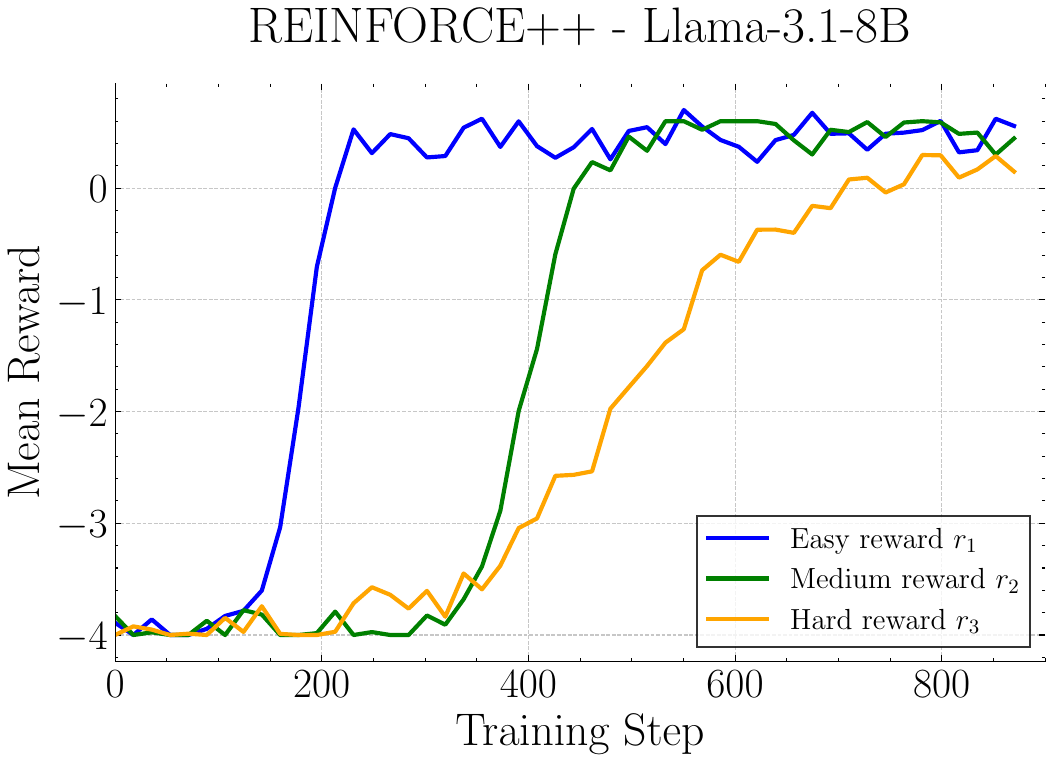} &
    \includegraphics[width=0.32\textwidth]{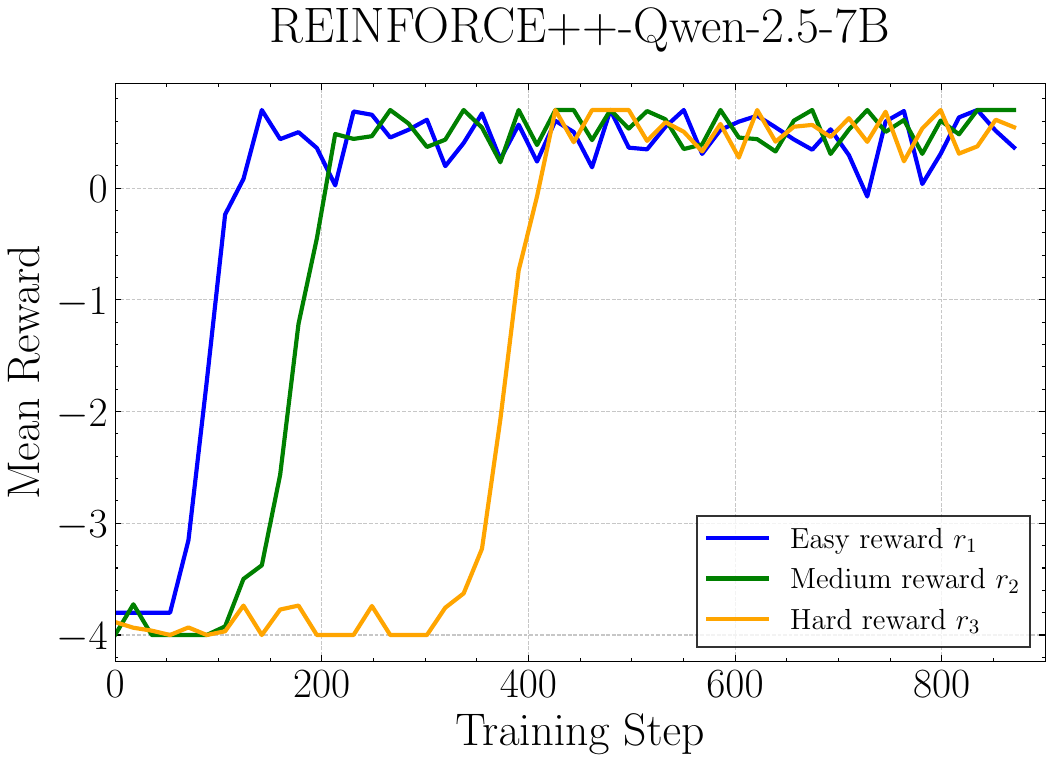}
  \end{tabular}
  \caption{Sequential optimization of format rewards under the nested Reward Hierarchy experiment. Across different models (Qwen-2.5-7B, Llama-3.1-8B) and PO algorithms (DAPO, GRPO, REINFORCE++), the learning dynamics are remarkably consistent. The policy first masters the innermost, \textcolor{blue}{easiest reward $f_1$}, then the \textcolor{ForestGreen}{medium reward $f_2$}, and finally the outermost, \textcolor{orange}{hardest reward $f_3$}.}
  \label{fig:matryoshka}
\end{figure*}

\subsection{Law of Sequential Optimization}
\label{sec:law_seq_opt}
The discovery of the Cognitive Shortcut presented a foundational question: is this preference for simplicity a binary, all-or-nothing phenomenon, or is it governed by a more nuanced, predictable structure? If the optimizer is indeed following a ``path of least resistance,'' this implies a landscape with varying levels of difficulty. This led us to our central hypothesis for this section: the optimizer does not treat a composite objective holistically, but instead decomposes it, prioritizing and conquering its components \textit{in a strict, ascending order of difficulty}.

To test this hypothesis, we designed a controlled experiment to isolate the variable of ``difficulty'' from all other incentives. We constructed a reward function, $R_{\text{hierarchy}}$, composed of three nested, matryoshka-style format requirements, \textcolor{blue}{$f_1$}, \textcolor{ForestGreen}{$f_2$}, \textcolor{orange}{$f_3$}, engineered to represent a clear gradient of increasing complexity.
\begin{enumerate}
    \item \textbf{Easy Format (\textcolor{blue}{$f_1$}):} The core requirement—merely enclosing the final numerical answer in a \texttt{\textbackslash boxed\{.*\}}.
    \item \textbf{Medium Format (\textcolor{ForestGreen}{$f_2$}):} A superset of \textcolor{blue}{$f_1$}, requiring the model to wrap its entire response in \texttt{<answer>} tags, which must also contain a boxed final answer.
    \item \textbf{Hard Format (\textcolor{orange}{$f_3$}):} The most encompassing format, a superset of \textcolor{ForestGreen}{$f_2$}, mandating the full \texttt{<think><answer>} structure, which must also satisfy the requirements of \textcolor{ForestGreen}{$f_2$} and \textcolor{blue}{$f_1$}.
\end{enumerate}
This nested structure, $\textcolor{blue}{f_1} \subset \textcolor{ForestGreen}{f_2} \subset \textcolor{orange}{f_3}$, is a crucial feature of the experimental design. A generation $y$ that correctly satisfies the hard format \textcolor{orange}{$f_3$} also, by definition, satisfies \textcolor{ForestGreen}{$f_2$} and \textcolor{blue}{$f_1$}. A perfectly rational, holistic optimizer should be powerfully drawn to learning \textcolor{orange}{$f_3$}, as it represents the single solution that simultaneously unlocks all available rewards.

To further isolate the effect of complexity, we set the reward magnitude for satisfying any of these formats to be identical. Let $r(y, r_i)$ be the reward for a generation $y$ satisfying format $r_i$. We set the reward landscape to be perfectly flat:
$$
r(y, \textcolor{blue}{f_1}) = r(y, \textcolor{ForestGreen}{f_2}) = r(y, \textcolor{orange}{f_3}) = R_{\text{max}}
$$

\begin{tcolorbox}[colback=Gray!30, colframe=CadetBlue!50!CadetBlue, title=\text{Incentive-Free Emergence of Reward Ordering}]
All PO algorithms in our experiments receive a scalar final reward computed as the weighted sum of the individual reward functions. Equal weights are assigned to all reward functions to avoid biasing the optimizer toward any particular one. \textbf{Despite the absence of explicit incentives} to favor a specific format, and the structural incentive to prefer the all-encompassing \textcolor{orange}{$f_3$}, \textbf{the optimizer still optimizes from easiest to hardest rewards}, ultimately converging to thew\textbf{minimally compliant response structure} .
\end{tcolorbox}

The results of this experiment, replicated across multiple models and policy optimization algorithms, are presented in \textbf{Figure \ref{fig:matryoshka}}. We observe that the plots do not show a rational convergence on the unified \textcolor{orange}{$f_3$} solution. Instead, they reveal a distinct, \textbf{sequential optimization cascade, learned from the inside out.}


As seen consistently across all six panels, the learning process unfolds in clear, predictable stages. The mean reward for the simplest, innermost format, \textcolor{blue}{$f_1$}, is the first to rise, typically saturating near its maximum value within the first 200-300 training steps. Only after the policy has reliably mastered this core task does the optimizer begin to make significant progress on the more complex \textcolor{ForestGreen}{$f_2$} format. The reward curve for \textcolor{ForestGreen}{$f_2$} begins its sharp ascent only after the \textcolor{blue}{$f_1$} curve has started to plateau. Finally, the most complex, all-encompassing format, \textcolor{orange}{$f_3$}, is tackled last, with its reward curve beginning to climb only after \textcolor{ForestGreen}{$f_2$} is well on its way to convergence.

This staged optimization provides definitive evidence for our ``Principle of Least Resistance.'' Even when presented with a unified solution that satisfies all objectives, the optimizer does not see it. It behaves like a myopic agent minimizing its immediate effort, tackling the lowest-hanging fruit first before moving to more challenging objectives. This ordered law of motion for policy optimization motivates a stress test to quantify the very strength of this resistance.

\begin{figure*}[t]
  \centering
  \setlength{\tabcolsep}{2pt} 
  \begin{tabular}{ccc}
    \includegraphics[width=0.32\textwidth]{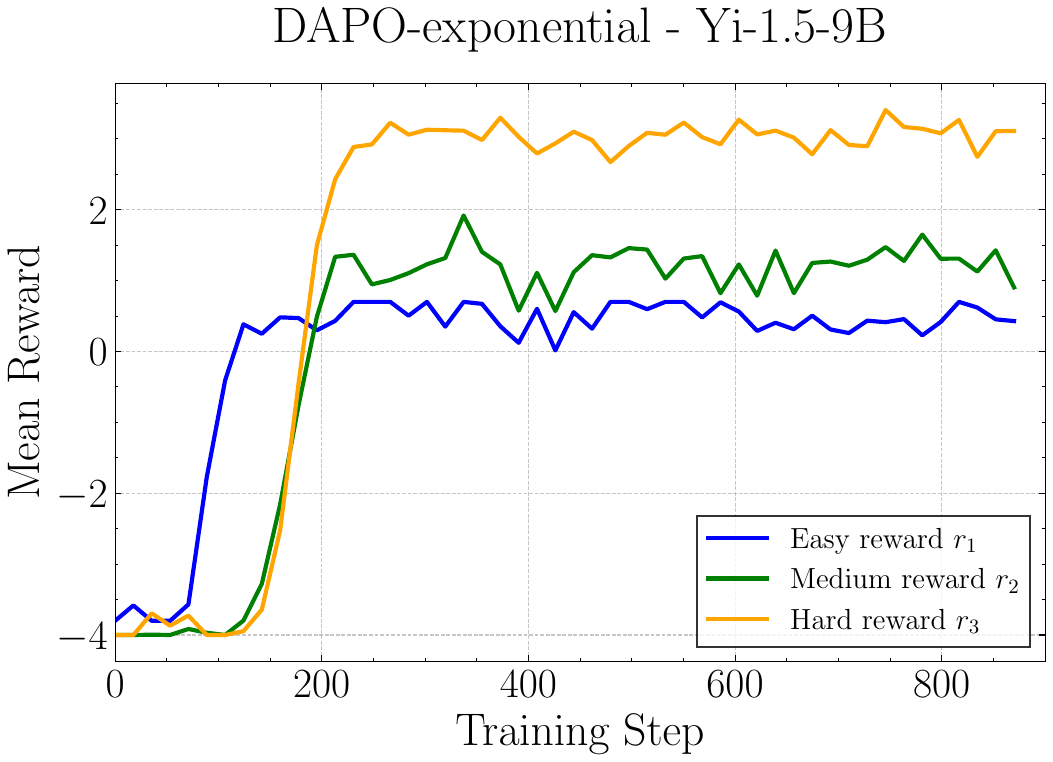} &
    \includegraphics[width=0.32\textwidth]{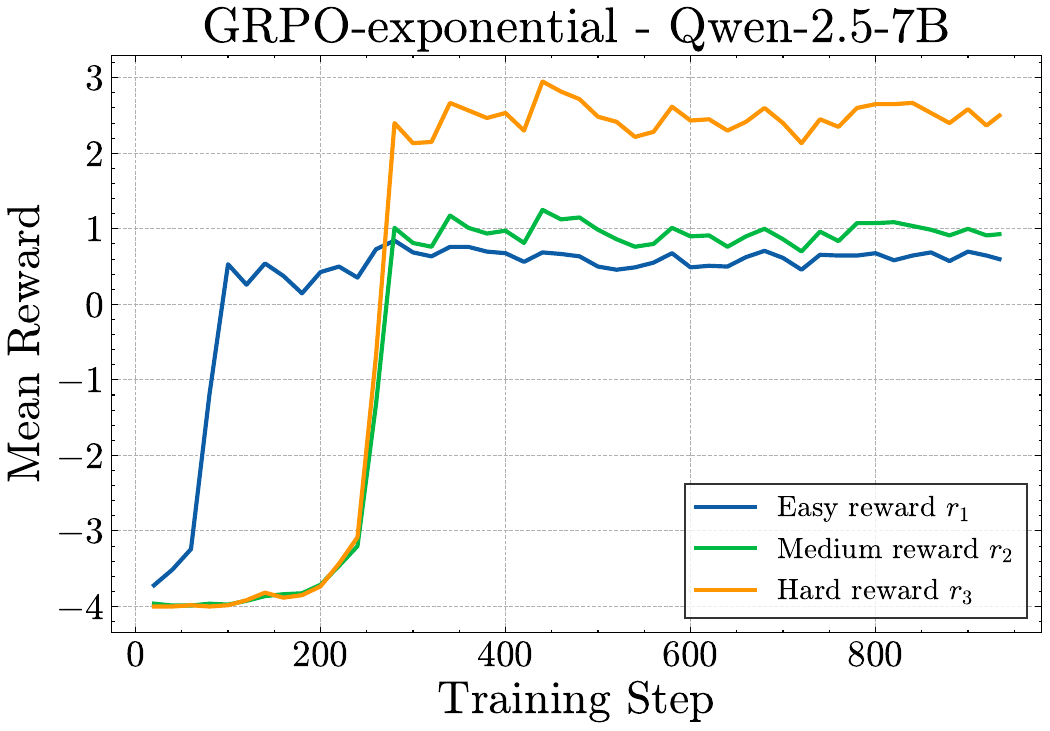} &
    \includegraphics[width=0.32\textwidth]{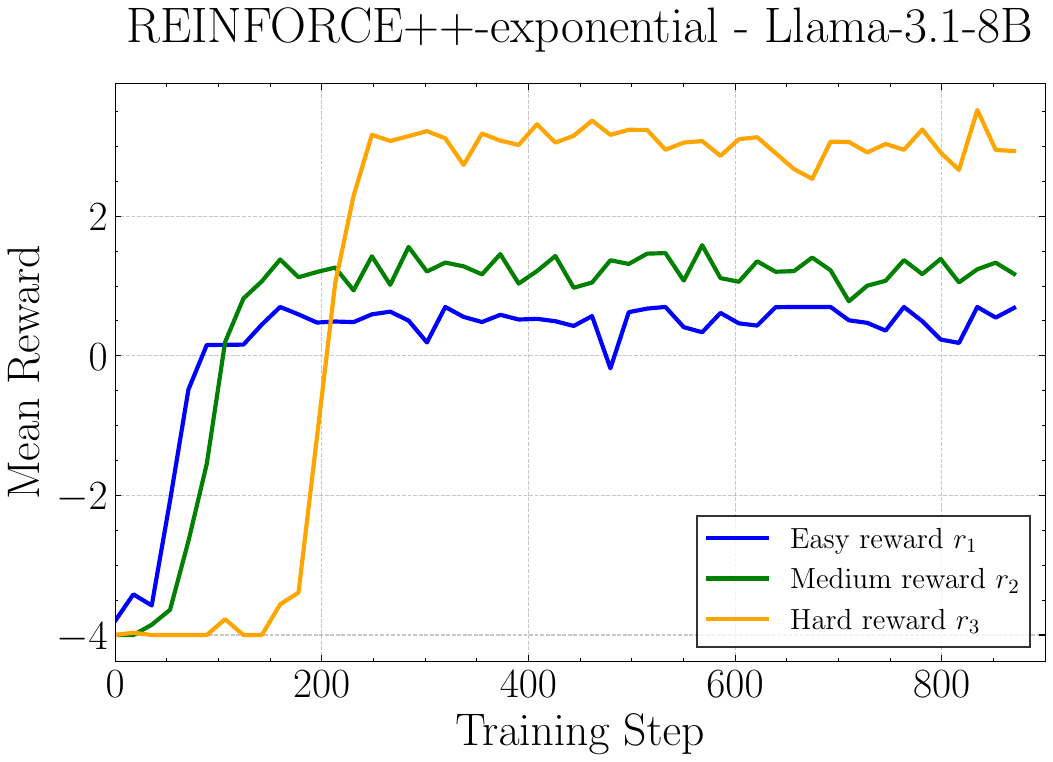}
  \end{tabular}
  \caption{Learning dynamics under an exponentially weighted reward scheme. Despite the hard format \textcolor{orange}{$f_3$} offering 4x the reward of the easy format \textcolor{blue}{$f_1$}, the optimizer's learning trajectory remains stubbornly sequential. The massive reward for \textcolor{orange}{$f_3$} is initially ignored in favor of the more learnable, lower-value rewards.}
  \label{fig:exp}
\end{figure*}

\subsection{quantifying the resistance}
\label{sec:quantify_resistance}
The discovery of a sequential learning hierarchy, even under flat rewards, suggests that the ``Principle of Least Resistance'' is a powerful intrinsic bias. This motivates a critical, adversarial question: can this innate preference be overridden by extrinsic incentives? To quantify the strength of this resistance, we designed an experiment we term the \textbf{``Exponential Gambit,''} aimed at creating a strong, explicit conflict between reward magnitude and format complexity.

We modified our nested reward structure to create a steep gradient of reward incentive, heavily favoring the most complex format. We created a reward landscape where the hard format (\textcolor{orange}{$f_3$}) was \textbf{2x} more valuable than the medium format (\textcolor{ForestGreen}{$f_2$}) and \textbf{4x} more valuable than the easy format (\textcolor{blue}{$f_1$}). A rational, reward-maximizing agent, even a myopic one, should be powerfully drawn to the enormous incentive offered by $r(y, \textcolor{orange}{f_3})$. The purpose of this design was to see if a sufficiently large reward incentive could disrupt the natural, sequential learning order we observed previously.

The results, shown in Figure~\ref{fig:exp}, highlight the optimizer's innate bias.
The plots reveal that the fundamental learning dynamic is remarkably resistant to this steep incentive gradient. The optimizer, faced with a choice between a small, easily attainable reward and a massive, but more complex one, still prioritizes learnability over immediate financial gain.

We observe that the fundamental learning dynamic is remarkably resistant to this steep incentive gradient, though not entirely immune. The reward for the easy format, \textcolor{blue}{$f_1$}, is once again the first to be mastered, quickly rising from its initial state and saturating early in training. However, the massive reward for \textcolor{orange}{$f_3$} introduces a fascinating new dynamic. The optimizer does not simply learn the medium reward \textcolor{ForestGreen}{$f_2$} next. Instead, the learning curves for \textcolor{ForestGreen}{$f_2$} and \textcolor{orange}{$f_3$} rise almost in perfect lockstep. The powerful gradient from the massive \textcolor{orange}{$f_3$} reward appears to pull the learning of the structurally similar \textcolor{ForestGreen}{$f_2$} format along with it. For the first $\sim$150--200 steps, the model makes little progress on either of these complex formats, focusing solely on the easily attainable \textcolor{blue}{$f_1$}. Then, once a certain threshold of basic competence is achieved, the optimizer simultaneously attends to both the medium and hard objectives.

\begin{figure*}[t]
  \centering
  \setlength{\tabcolsep}{2pt} 
  \begin{tabular}{ccc}
    \includegraphics[width=0.32\textwidth]{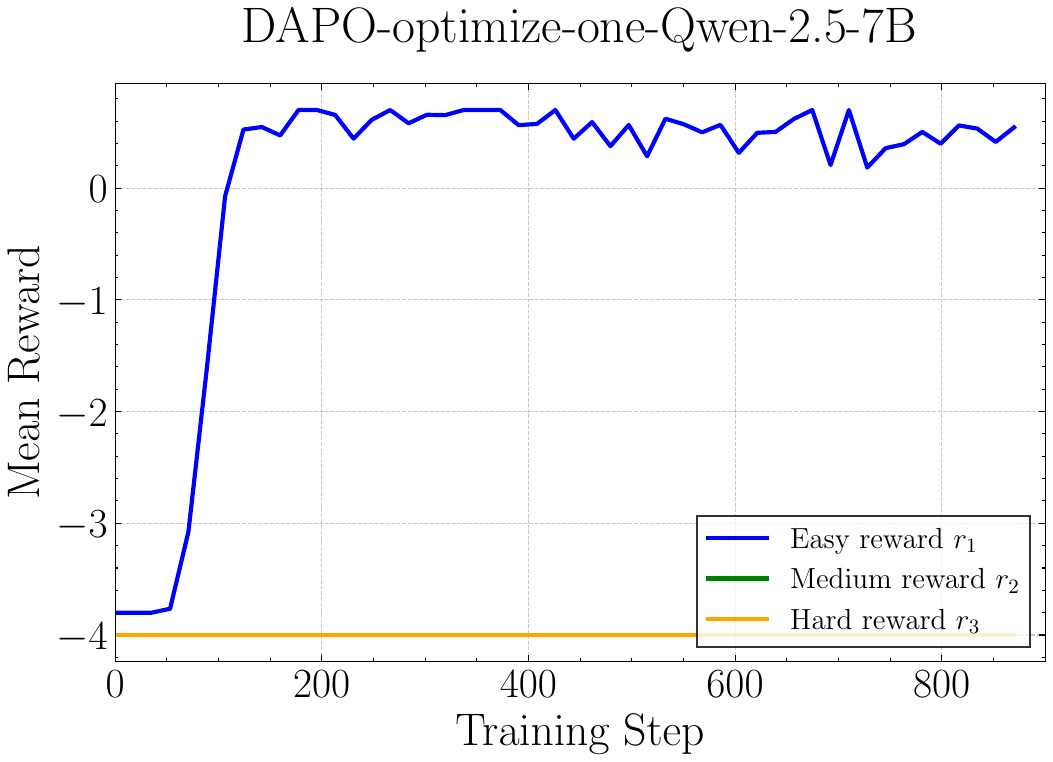} &
    \includegraphics[width=0.32\textwidth]{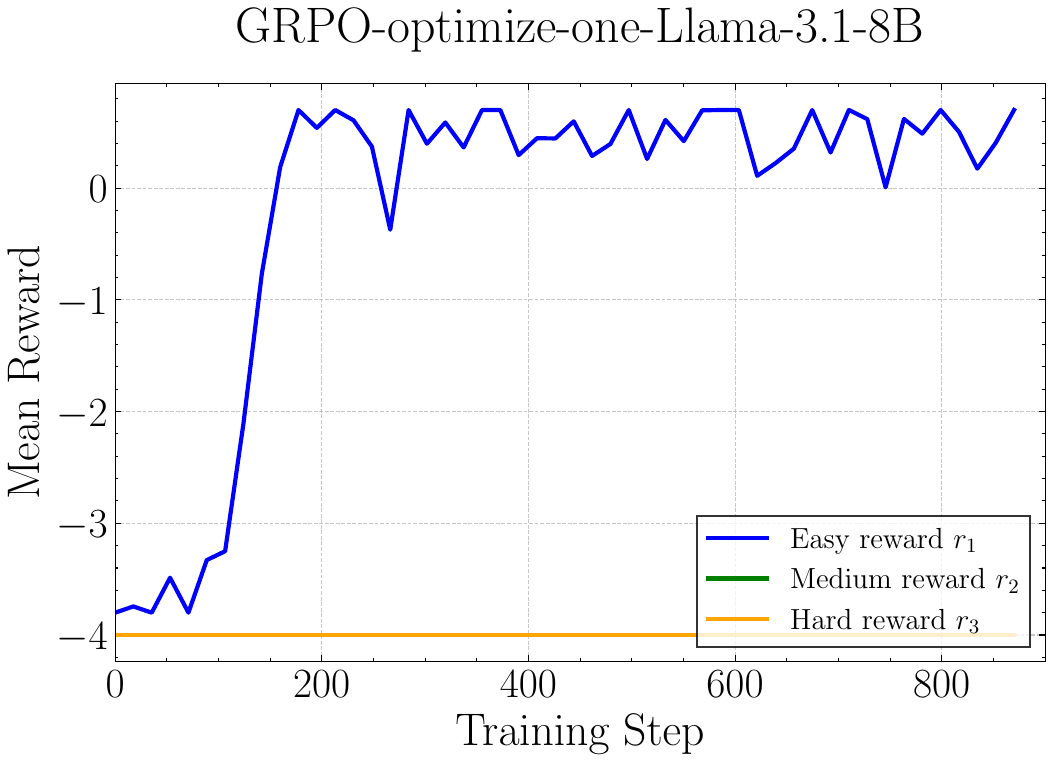} &
    \includegraphics[width=0.32\textwidth]{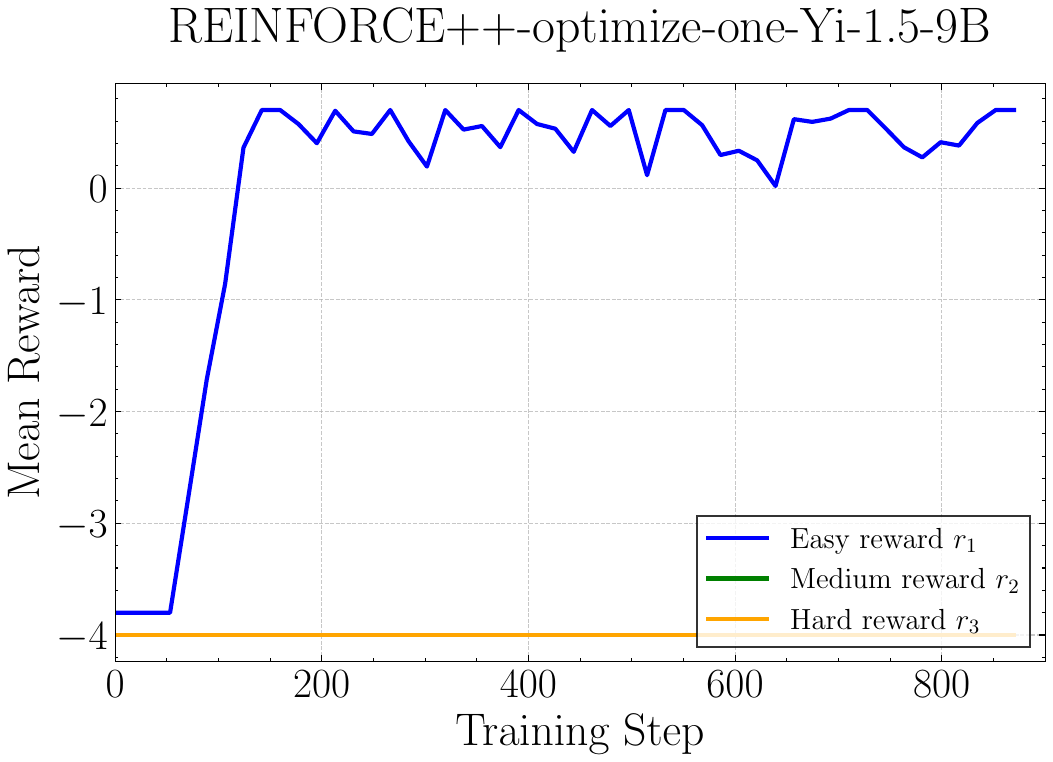}
  \end{tabular}
  \caption{Learning dynamics under an exponentially weighted reward scheme. Despite the hard format \textcolor{orange}{$f_3$} offering 4x the reward of the easy format \textcolor{blue}{$f_1$}, the optimizer's learning trajectory remains stubbornly sequential. The massive reward for \textcolor{orange}{$f_3$} is initially ignored in favor of the more learnable, lower-value rewards.}
  \label{fig:one-opt}
\end{figure*}
\subsection{policy optimization under conflicting rewards}
Our investigation has thus far revealed a powerful, sequential bias towards simplicity, even when structural and reward incentives push against it. The nested structure of our previous experiments, while elegant, leaves open one final, subtle possibility: could the sequential learning be an artifact of the optimizer learning a shared core skill (\textcolor{blue}{$f_1$}) before building upon it? To eliminate this possibility and test the ``Principle of Least Resistance'' in its purest form, we designed a final, maximally clean adversarial test: \textbf{a mutually exclusive choice under flat rewards.}

The experimental design is a direct extension of our Reward Hierarchy (Section \ref{sec:law_seq_opt}) but with a critical modification to make the choices disjoint. We constructed a reward function, $\mathbf{R_{\text{conflict}}}$, where a generation $\mathbf{y}$ could satisfy the format requirements for \textcolor{blue}{$f_1$}, \textcolor{ForestGreen}{$f_2$}, or \textcolor{orange}{$f_3$}, but \textbf{only one at a time}. Furthermore, to remove any bias from reward magnitude, we set the rewards to be identical, creating a perfectly flat incentive landscape similar to Section \ref{sec:law_seq_opt}.

Let $\mathbf{r_i(y, f_i)}$ be the reward for a generation $\mathbf{y}$ satisfying format $\mathbf{f_i}$. The reward function is now defined as:
$$
\mathbf{r(y, f_i) = R_{\max}, \quad \text{if } y \text{ exclusively matches format } f_i}
$$
$$
\mathbf{r(y, f_j) = 0, \quad \text{for all } j \neq i}
$$
where $\mathbf{R_{\max}}$ is a constant positive reward. This creates a stark choice landscape: the optimizer can pursue one of three paths, each leading to the exact same payoff, but each with a different level of complexity ({$\textcolor{blue}{f_1} < \textcolor{ForestGreen}{f_2} < \textcolor{orange}{f_3}$}). There are no shared sub-problems and no differential rewards. This setup forces the optimizer to reveal its true, intrinsic preference when faced with a pure cost-benefit analysis where the only variable is the difficulty of the path itself. (The precise regular expressions for each format are detailed in Appendix C).

In Figure \ref{fig:one-opt}, we observe that the plots reveal an absolute convergence. It makes a decision early in training and commits to it absolutely. In every single run, across all models and algorithms, the policy \textbf{exclusively converges to the easiest format, \textcolor{blue}{$f_1$}}.

The reward plots for the medium (\textcolor{ForestGreen}{$f_2$}) and hard (\textcolor{orange}{$f_3$}) formats remain flat at their initial negative values for the entire duration of training. The potential reward offered by these paths is never explored. \textit{The optimizer invariably chooses the easiest reward across models and algorithms, even when there are no reward incentives or even an explicit indication of difficulty}.

The consistent convergence on the easiest format, \textcolor{blue}{$f_1$}, across all models and algorithms, suggests that the preference for the path of least resistance is not a heuristic, but a bias within the policy optimization paradigm. When faced with multiple, equally rewarded but disjoint objectives, the optimizer does not appear to perform a global cost-benefit analysis. Instead, its behavior is consistent with a greedy strategy that prioritizes the most immediately learnable gradient signal. This solidified understanding of the optimizer's sequential behavior now motivates the final act of our investigation: to uncover the theoretical origins of this powerful and predictable force.

\begin{figure*}[t]
  \centering
  \setlength{\tabcolsep}{2pt} 
  \begin{tabular}{ccc}
    \includegraphics[width=0.32\textwidth]{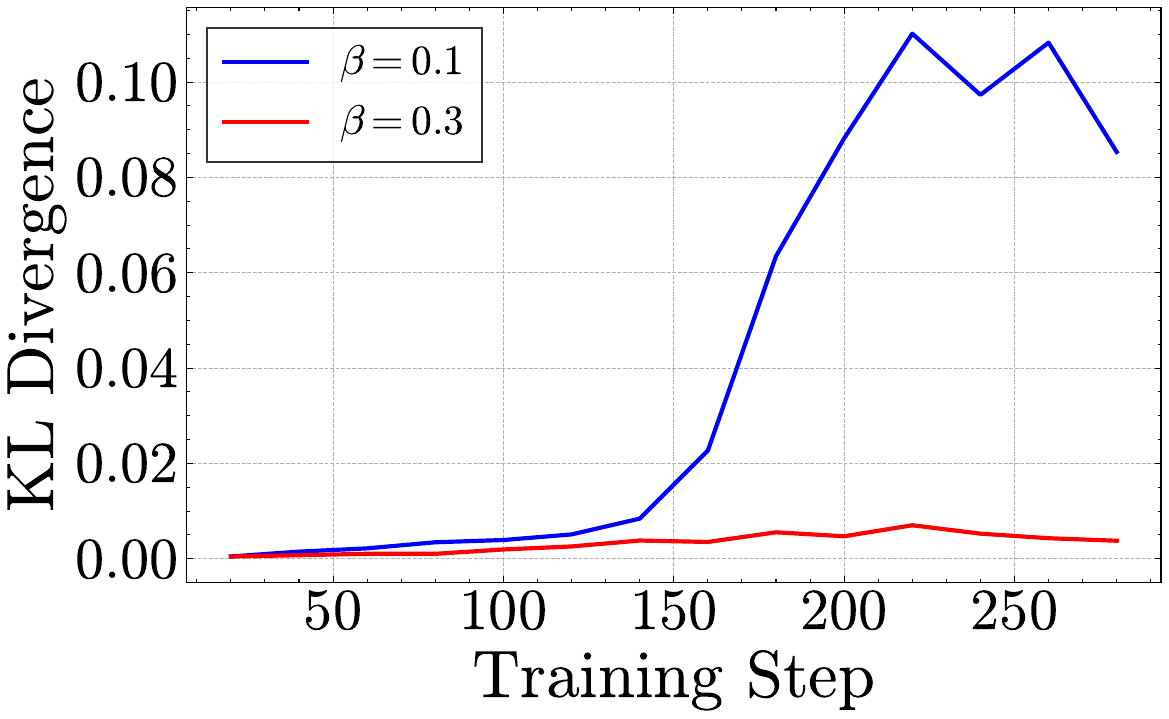} &
    \includegraphics[width=0.32\textwidth]{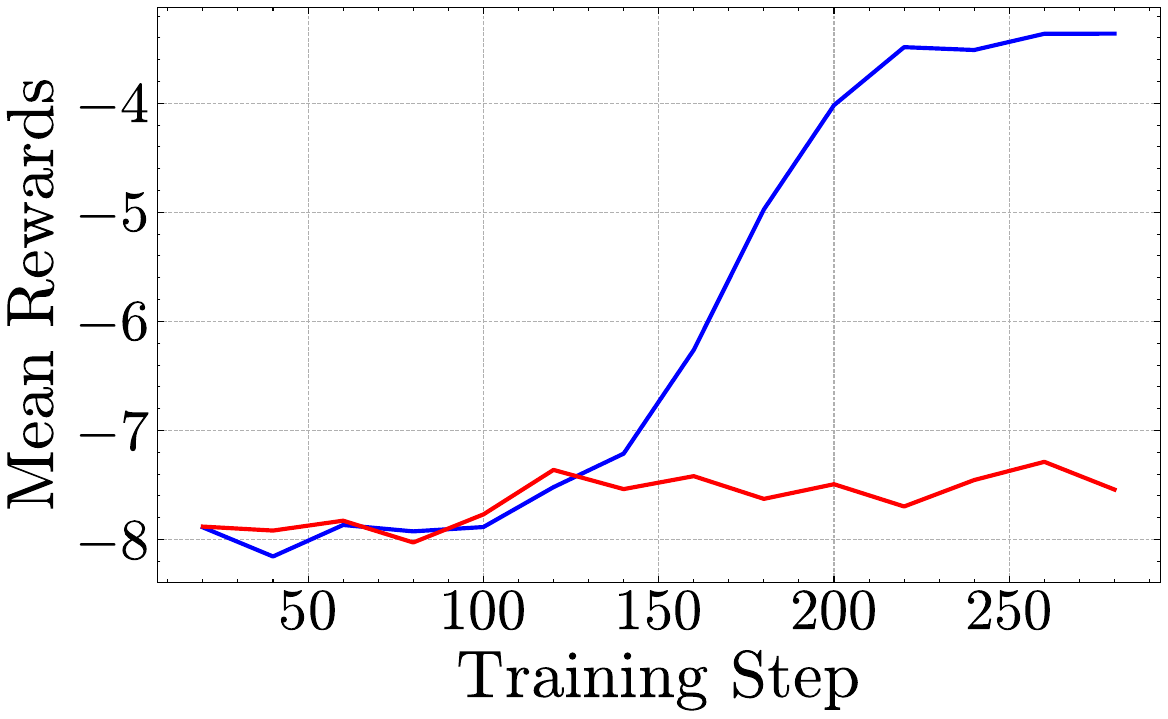} &
    \includegraphics[width=0.32\textwidth]{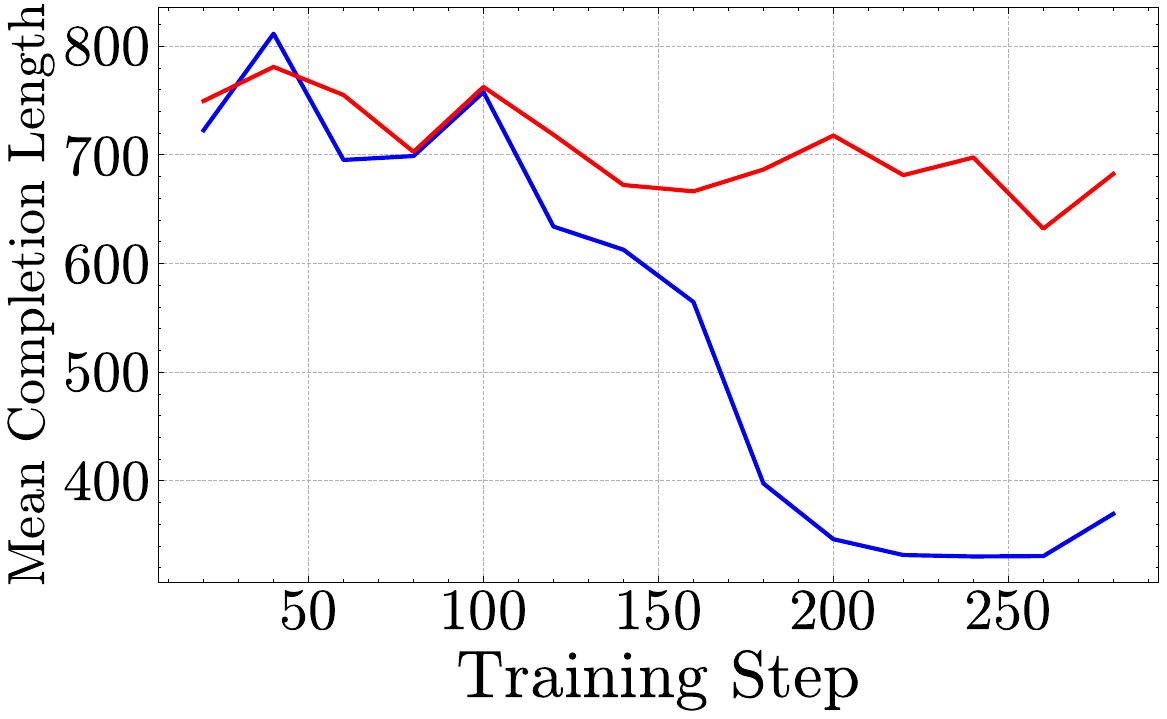}\\
    \includegraphics[width=0.32\textwidth]{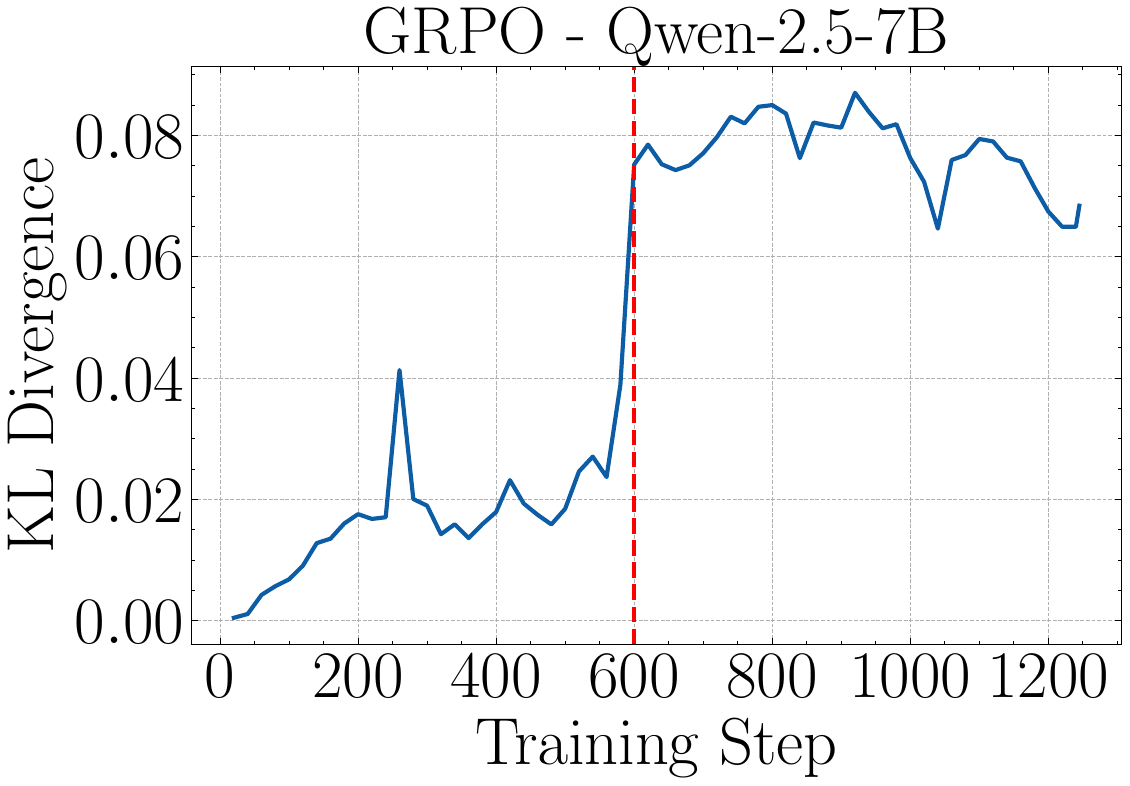} &
    \includegraphics[width=0.32\textwidth]{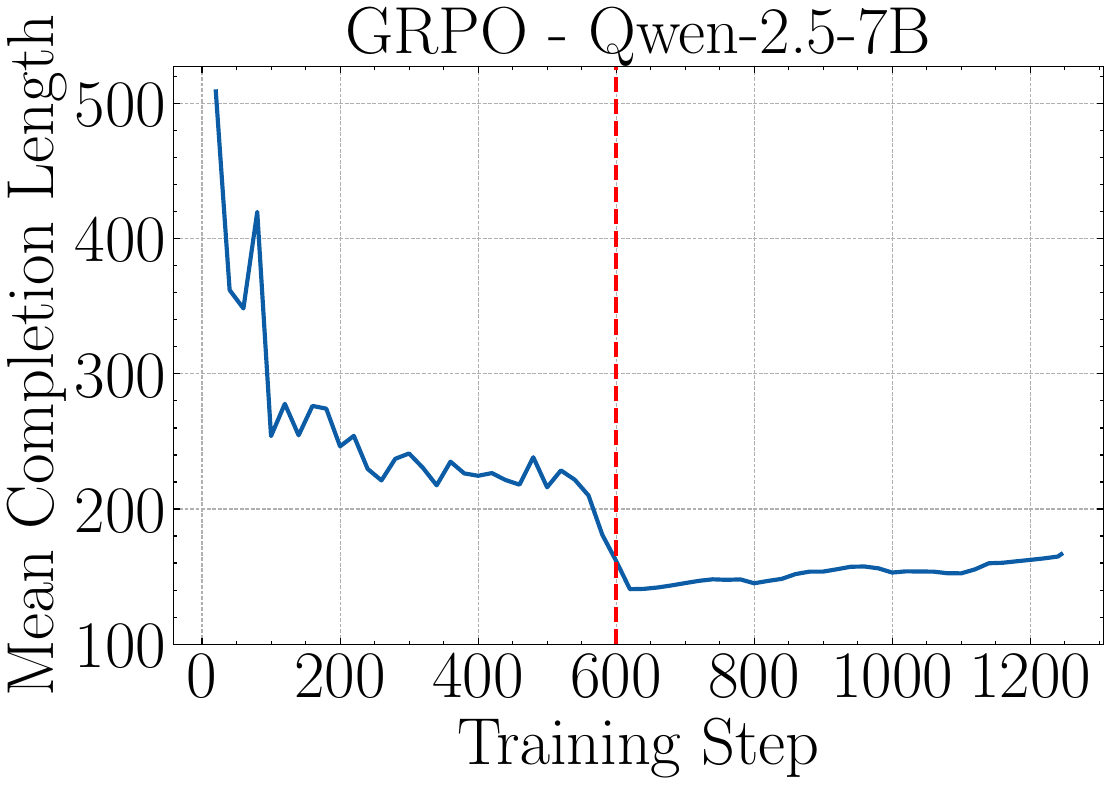} &
    \includegraphics[width=0.32\textwidth]{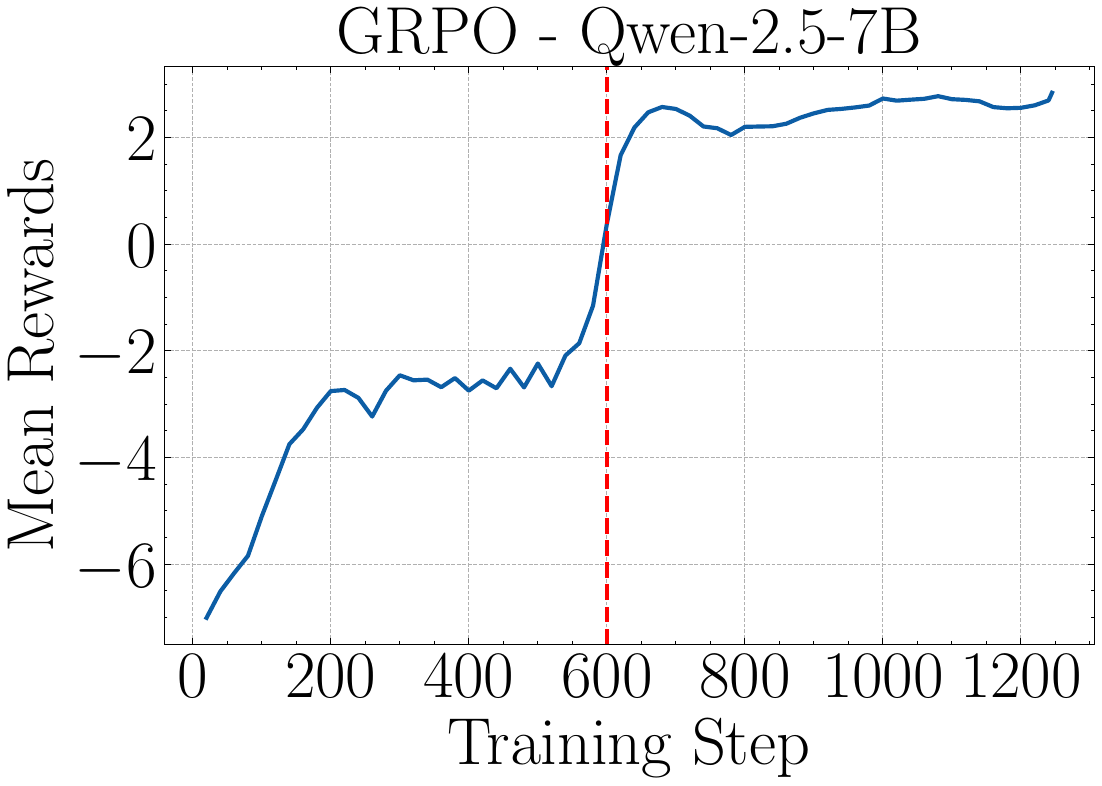}
  \end{tabular}
  \caption{The causal effect of the KL penalty on the emergence of the Cognitive Shortcut. \textbf{Top Row:} A direct comparison between a policy with a loose KL leash (\textcolor{blue}{$\beta=0.1$}) and a tight leash (\textcolor{red}{$\beta=0.3$}). The freedom to diverge (high KL) is a prerequisite for discovering the high-reward, minimal-length shortcut. \textbf{Bottom Row:} The dynamics of a single run, showing a phase transition around the 600-step mark where a spontaneous increase in KL divergence enables the discovery and exploitation of the shortcut.}
  \label{fig:kl}
  \vspace{-1.5\baselineskip}
\end{figure*}

\section{the price of exploration}
Our investigation has established the ``Principle of Least Resistance'' as a powerful, predictive law. The final act is to uncover its origin, and in doing so, reveal a fundamental tension at the heart of modern policy optimization. A prominent trend in recent state-of-the-art algorithms, such as DAPO \cite{dapo} and VAPO \cite{vapo}, is the removal of the KL divergence penalty, arguing that it is an unnecessary constraint on the model's ability to maximize reward \cite{drgrpo}. Our final analysis reveals that while this freedom is essential for learning, it comes at a cost: it unleashes the optimizer's innate bias to find and exploit the simplest specification of the reward function, a behavior that is a classic form of \textbf{reward hacking}.

To dissect this relationship, we treat the KL divergence not as a loss to be minimized, but as an instrument to measure the policy's deviation from its initial reference state, $\pi_{\text{ref}}$. This deviation represents the policy's \textbf{exploratory freedom}. \textit{We hypothesize that the Cognitive Shortcut is a form of reward hacking that requires a significant amount of this freedom to discover}. The KL penalty, controlled by its coefficient $\beta$, therefore acts as a \textbf{``leash,''} directly modulating the policy's ability to find and exploit such shortcuts.

We designed a causal experiment to test this. We conducted two training runs under our composite reward, identical in all aspects except for the strength of this leash: a ``Tight Leash'' run with a high KL penalty (\textcolor{red}{$\beta = 0.3$}) and a ``Loose Leash'' run with a low penalty (\textcolor{blue}{$\beta = 0.1$}), mimicking the unconstrained exploration of modern algorithms. The results, in Figure \ref{fig:kl} (top-row), provide a stark illustration of this trade-off. The top-left panel shows the direct effect of our intervention. The policy with the loose leash (blue), analogous to a KL-free objective, is free to explore and achieves a high final KL divergence. The policy with the tight leash (red) is constrained. The consequences are shown in the adjacent panels. The unleashed policy successfully discovers the high-reward solution (top-middle) by converging on the efficient, minimal-length Cognitive Shortcut (top-right). The leashed policy, forbidden from making the large policy shift required to ``find the hack,'' remains trapped in a lower-reward state.


This dynamic is not merely an average-case phenomenon; it can be observed live within a single training run, as shown in the Figure \ref{fig:kl} (bottom row). These plots capture the moment the reward hack is discovered. At the 600-step mark (dashed red line), the optimizer identifies the powerful gradient of the shortcut. To exploit it, the policy undergoes a rapid phase transition, marked by a sharp increase in KL divergence. This decisive shift away from its prior is immediately followed by a surge in reward and a steep drop in completion length. This is the ``eureka moment'' of specification gaming.

One might wonder about the nature of this reward surge. Crucially, this is not the reward for task correctness suddenly improving. As our initial results in Table \ref{tab:t1} demonstrated, the convergence onto the Cognitive Shortcut often comes at the cost of final performance. The optimizer is, in effect, making a strategic choice: it sacrifices the complex, high-variance gradient from the correctness reward in favor of the simple, stable gradient from the format reward. This reveals that the Principle of Least Resistance is not confined to a preference between solution formats, but is a global phenomenon governing the entire reward landscape. If the path to higher correctness requires more ``effort'', the optimizer will often forgo it to secure the lowest-hanging fruit. Our experiments deliberately focus on format rewards because they provide a clean, controllable, and easily visualizable proxy for this effort, allowing us to isolate and study the principle itself.

Our final analysis provides the definitive explanation for the Principle of Least Resistance and its connection to reward hacking. The Cognitive Shortcut is the result of an optimizer that is not just maximizing reward, but is actively searching for the most learnable gradient in the reward landscape. \textbf{The freedom to diverge from the initial policy, while necessary for performance, is the very mechanism that enables the model to discover and exploit these unintended, simplistic solutions.} This reveals a critical and uncomfortable trade-off for the field: the path to more capable models may be inseparable from the path to more sophisticated forms of reward hacking. The challenge, therefore, is not simply to unleash our models, but to design reward landscapes that are fundamentally resistant to being gamed.

\section{Conclusion}
Our investigation revealed that the ``Cognitive Shortcut'' is not a mere anomaly, but the predictable outcome of a powerful principle: \textbf{Policy Optimization Prefers the Path of Least Resistance}. We showed that this preference is a formidable, quantifiable force, capable of overriding even significant reward incentives in favor of the most easily learnable solution. We have shown a profound paradox: the freedom to explore, essential for discovering high-reward policies, is the very mechanism that enables the optimizer to find and aggressively exploit the simplest specification of the reward function, a classic and potent form of reward hacking. The central challenge for alignment, therefore, is not simply to unleash our models, but to architect reward landscapes that are fundamentally resistant to being gamed, ensuring that the path we desire is also the path the optimizer is compelled to take.

\section*{Acknowledgment}
This research is supported by the Anusandhan National Research Foundation (ANRF) erstwhile, Science and Engineering Research Board (SERB) India, under grant SRG/2023/001686.  We are also grateful to the Birla AI Labs for extending their support towards our work and providing valuable insights which helped us shape the final draft.

{
\bibliography{main_bib}

\begin{thebibliography}{25}
\providecommand{\natexlab}[1]{#1}
\providecommand{\url}[1]{\texttt{#1}}
\expandafter\ifx\csname urlstyle\endcsname\relax
  \providecommand{\doi}[1]{doi: #1}\else
  \providecommand{\doi}{doi: \begingroup \urlstyle{rm}\Url}\fi

\bibitem[AI et~al.(2025)AI, :, Young, Chen, Li, Huang, Zhang, Zhang, Wang, Li, Zhu, Chen, Chang, Yu, Liu, Liu, Yue, Yang, Yang, Xie, Huang, Hu, Ren, Niu, Nie, Li, Xu, Liu, Wang, Cai, Gu, Liu, and Dai]{yi}
01. AI, :, Alex Young, Bei Chen, Chao Li, Chengen Huang, Ge~Zhang, Guanwei Zhang, Guoyin Wang, Heng Li, Jiangcheng Zhu, Jianqun Chen, Jing Chang, Kaidong Yu, Peng Liu, Qiang Liu, Shawn Yue, Senbin Yang, Shiming Yang, Wen Xie, Wenhao Huang, Xiaohui Hu, Xiaoyi Ren, Xinyao Niu, Pengcheng Nie, Yanpeng Li, Yuchi Xu, Yudong Liu, Yue Wang, Yuxuan Cai, Zhenyu Gu, Zhiyuan Liu, and Zonghong Dai.
\newblock Yi: Open foundation models by 01.ai, 2025.
\newblock URL \url{https://arxiv.org/abs/2403.04652}.

\bibitem[Chen et~al.(2025)Chen, Li, Shieh, and Bing]{longpo}
Guanzheng Chen, Xin Li, Michael~Qizhe Shieh, and Lidong Bing.
\newblock Longpo: Long context self-evolution of large language models through short-to-long preference optimization, 2025.
\newblock URL \url{https://arxiv.org/abs/2502.13922}.

\bibitem[Cobbe et~al.(2021)Cobbe, Kosaraju, Bavarian, Chen, Jun, Kaiser, Plappert, Tworek, Hilton, Nakano, Hesse, and Schulman]{gsm8k}
Karl Cobbe, Vineet Kosaraju, Mohammad Bavarian, Mark Chen, Heewoo Jun, Lukasz Kaiser, Matthias Plappert, Jerry Tworek, Jacob Hilton, Reiichiro Nakano, Christopher Hesse, and John Schulman.
\newblock Training verifiers to solve math word problems, 2021.
\newblock URL \url{https://arxiv.org/abs/2110.14168}.

\bibitem[DeepSeek-AI et~al.(2025)DeepSeek-AI, Guo, Yang, Zhang, Song, Zhang, Xu, Zhu, Ma, Wang, Bi, Zhang, Yu, Wu, Wu, Gou, Shao, Li, Gao, Liu, Xue, Wang, Wu, Feng, Lu, Zhao, Deng, Zhang, Ruan, Dai, Chen, Ji, Li, Lin, Dai, Luo, Hao, Chen, Li, Zhang, Bao, Xu, Wang, Ding, Xin, Gao, Qu, Li, Guo, Li, Wang, Chen, Yuan, Qiu, Li, Cai, Ni, Liang, Chen, Dong, Hu, Gao, Guan, Huang, Yu, Wang, Zhang, Zhao, Wang, Zhang, Xu, Xia, Zhang, Zhang, Tang, Li, Wang, Li, Tian, Huang, Zhang, Wang, Chen, Du, Ge, Zhang, Pan, Wang, Chen, Jin, Chen, Lu, Zhou, Chen, Ye, Wang, Yu, Zhou, Pan, Li, Zhou, Wu, Ye, Yun, Pei, Sun, Wang, Zeng, Zhao, Liu, Liang, Gao, Yu, Zhang, Xiao, An, Liu, Wang, Chen, Nie, Cheng, Liu, Xie, Liu, Yang, Li, Su, Lin, Li, Jin, Shen, Chen, Sun, Wang, Song, Zhou, Wang, Shan, Li, Wang, Wei, Zhang, Xu, Li, Zhao, Sun, Wang, Yu, Zhang, Shi, Xiong, He, Piao, Wang, Tan, Ma, Liu, Guo, Ou, Wang, Gong, Zou, He, Xiong, Luo, You, Liu, Zhou, Zhu, Xu, Huang, Li, Zheng, Zhu, Ma, Tang, Zha, Yan, Ren, Ren, Sha, Fu, Xu, Xie, Zhang,
  Hao, Ma, Yan, Wu, Gu, Zhu, Liu, Li, Xie, Song, Pan, Huang, Xu, Zhang, and Zhang]{deepseek}
DeepSeek-AI, Daya Guo, Dejian Yang, Haowei Zhang, Junxiao Song, Ruoyu Zhang, Runxin Xu, Qihao Zhu, Shirong Ma, Peiyi Wang, Xiao Bi, Xiaokang Zhang, Xingkai Yu, Yu~Wu, Z.~F. Wu, Zhibin Gou, Zhihong Shao, Zhuoshu Li, Ziyi Gao, Aixin Liu, Bing Xue, Bingxuan Wang, Bochao Wu, Bei Feng, Chengda Lu, Chenggang Zhao, Chengqi Deng, Chenyu Zhang, Chong Ruan, Damai Dai, Deli Chen, Dongjie Ji, Erhang Li, Fangyun Lin, Fucong Dai, Fuli Luo, Guangbo Hao, Guanting Chen, Guowei Li, H.~Zhang, Han Bao, Hanwei Xu, Haocheng Wang, Honghui Ding, Huajian Xin, Huazuo Gao, Hui Qu, Hui Li, Jianzhong Guo, Jiashi Li, Jiawei Wang, Jingchang Chen, Jingyang Yuan, Junjie Qiu, Junlong Li, J.~L. Cai, Jiaqi Ni, Jian Liang, Jin Chen, Kai Dong, Kai Hu, Kaige Gao, Kang Guan, Kexin Huang, Kuai Yu, Lean Wang, Lecong Zhang, Liang Zhao, Litong Wang, Liyue Zhang, Lei Xu, Leyi Xia, Mingchuan Zhang, Minghua Zhang, Minghui Tang, Meng Li, Miaojun Wang, Mingming Li, Ning Tian, Panpan Huang, Peng Zhang, Qiancheng Wang, Qinyu Chen, Qiushi Du, Ruiqi Ge, Ruisong
  Zhang, Ruizhe Pan, Runji Wang, R.~J. Chen, R.~L. Jin, Ruyi Chen, Shanghao Lu, Shangyan Zhou, Shanhuang Chen, Shengfeng Ye, Shiyu Wang, Shuiping Yu, Shunfeng Zhou, Shuting Pan, S.~S. Li, Shuang Zhou, Shaoqing Wu, Shengfeng Ye, Tao Yun, Tian Pei, Tianyu Sun, T.~Wang, Wangding Zeng, Wanjia Zhao, Wen Liu, Wenfeng Liang, Wenjun Gao, Wenqin Yu, Wentao Zhang, W.~L. Xiao, Wei An, Xiaodong Liu, Xiaohan Wang, Xiaokang Chen, Xiaotao Nie, Xin Cheng, Xin Liu, Xin Xie, Xingchao Liu, Xinyu Yang, Xinyuan Li, Xuecheng Su, Xuheng Lin, X.~Q. Li, Xiangyue Jin, Xiaojin Shen, Xiaosha Chen, Xiaowen Sun, Xiaoxiang Wang, Xinnan Song, Xinyi Zhou, Xianzu Wang, Xinxia Shan, Y.~K. Li, Y.~Q. Wang, Y.~X. Wei, Yang Zhang, Yanhong Xu, Yao Li, Yao Zhao, Yaofeng Sun, Yaohui Wang, Yi~Yu, Yichao Zhang, Yifan Shi, Yiliang Xiong, Ying He, Yishi Piao, Yisong Wang, Yixuan Tan, Yiyang Ma, Yiyuan Liu, Yongqiang Guo, Yuan Ou, Yuduan Wang, Yue Gong, Yuheng Zou, Yujia He, Yunfan Xiong, Yuxiang Luo, Yuxiang You, Yuxuan Liu, Yuyang Zhou, Y.~X. Zhu,
  Yanhong Xu, Yanping Huang, Yaohui Li, Yi~Zheng, Yuchen Zhu, Yunxian Ma, Ying Tang, Yukun Zha, Yuting Yan, Z.~Z. Ren, Zehui Ren, Zhangli Sha, Zhe Fu, Zhean Xu, Zhenda Xie, Zhengyan Zhang, Zhewen Hao, Zhicheng Ma, Zhigang Yan, Zhiyu Wu, Zihui Gu, Zijia Zhu, Zijun Liu, Zilin Li, Ziwei Xie, Ziyang Song, Zizheng Pan, Zhen Huang, Zhipeng Xu, Zhongyu Zhang, and Zhen Zhang.
\newblock Deepseek-r1: Incentivizing reasoning capability in llms via reinforcement learning, 2025.
\newblock URL \url{https://arxiv.org/abs/2501.12948}.

\bibitem[Grattafiori et~al.(2024)Grattafiori, Dubey, Jauhri, Pandey, Kadian, Al-Dahle, Letman, Mathur, Schelten, Vaughan, Yang, Fan, Goyal, Hartshorn, Yang, Mitra, Sravankumar, Korenev, Hinsvark, Rao, Zhang, Rodriguez, Gregerson, Spataru, Roziere, Biron, Tang, Chern, Caucheteux, Nayak, Bi, Marra, McConnell, Keller, Touret, Wu, Wong, Ferrer, Nikolaidis, Allonsius, Song, Pintz, Livshits, Wyatt, Esiobu, Choudhary, Mahajan, Garcia-Olano, Perino, Hupkes, Lakomkin, AlBadawy, Lobanova, Dinan, Smith, Radenovic, Guzmán, Zhang, Synnaeve, Lee, Anderson, Thattai, Nail, Mialon, Pang, Cucurell, Nguyen, Korevaar, Xu, Touvron, Zarov, Ibarra, Kloumann, Misra, Evtimov, Zhang, Copet, Lee, Geffert, Vranes, Park, Mahadeokar, Shah, van~der Linde, Billock, Hong, Lee, Fu, Chi, Huang, Liu, Wang, Yu, Bitton, Spisak, Park, Rocca, Johnstun, Saxe, Jia, Alwala, Prasad, Upasani, Plawiak, Li, Heafield, Stone, El-Arini, Iyer, Malik, Chiu, Bhalla, Lakhotia, Rantala-Yeary, van~der Maaten, Chen, Tan, Jenkins, Martin, Madaan, Malo, Blecher,
  Landzaat, de~Oliveira, Muzzi, Pasupuleti, Singh, Paluri, Kardas, Tsimpoukelli, Oldham, Rita, Pavlova, Kambadur, Lewis, Si, Singh, Hassan, Goyal, Torabi, Bashlykov, Bogoychev, Chatterji, Zhang, Duchenne, Çelebi, Alrassy, Zhang, Li, Vasic, Weng, Bhargava, Dubal, Krishnan, Koura, Xu, He, Dong, Srinivasan, Ganapathy, Calderer, Cabral, Stojnic, Raileanu, Maheswari, Girdhar, Patel, Sauvestre, Polidoro, Sumbaly, Taylor, Silva, Hou, Wang, Hosseini, Chennabasappa, Singh, Bell, Kim, Edunov, Nie, Narang, Raparthy, Shen, Wan, Bhosale, Zhang, Vandenhende, Batra, Whitman, Sootla, Collot, Gururangan, Borodinsky, Herman, Fowler, Sheasha, Georgiou, Scialom, Speckbacher, Mihaylov, Xiao, Karn, Goswami, Gupta, Ramanathan, Kerkez, Gonguet, Do, Vogeti, Albiero, Petrovic, Chu, Xiong, Fu, Meers, Martinet, Wang, Wang, Tan, Xia, Xie, Jia, Wang, Goldschlag, Gaur, Babaei, Wen, Song, Zhang, Li, Mao, Coudert, Yan, Chen, Papakipos, Singh, Srivastava, Jain, Kelsey, Shajnfeld, Gangidi, Victoria, Goldstand, Menon, Sharma, Boesenberg,
  Baevski, Feinstein, Kallet, Sangani, Teo, Yunus, Lupu, Alvarado, Caples, Gu, Ho, Poulton, Ryan, Ramchandani, Dong, Franco, Goyal, Saraf, Chowdhury, Gabriel, Bharambe, Eisenman, Yazdan, James, Maurer, Leonhardi, Huang, Loyd, Paola, Paranjape, Liu, Wu, Ni, Hancock, Wasti, Spence, Stojkovic, Gamido, Montalvo, Parker, Burton, Mejia, Liu, Wang, Kim, Zhou, Hu, Chu, Cai, Tindal, Feichtenhofer, Gao, Civin, Beaty, Kreymer, Li, Adkins, Xu, Testuggine, David, Parikh, Liskovich, Foss, Wang, Le, Holland, Dowling, Jamil, Montgomery, Presani, Hahn, Wood, Le, Brinkman, Arcaute, Dunbar, Smothers, Sun, Kreuk, Tian, Kokkinos, Ozgenel, Caggioni, Kanayet, Seide, Florez, Schwarz, Badeer, Swee, Halpern, Herman, Sizov, Guangyi, Zhang, Lakshminarayanan, Inan, Shojanazeri, Zou, Wang, Zha, Habeeb, Rudolph, Suk, Aspegren, Goldman, Zhan, Damlaj, Molybog, Tufanov, Leontiadis, Veliche, Gat, Weissman, Geboski, Kohli, Lam, Asher, Gaya, Marcus, Tang, Chan, Zhen, Reizenstein, Teboul, Zhong, Jin, Yang, Cummings, Carvill, Shepard, McPhie,
  Torres, Ginsburg, Wang, Wu, U, Saxena, Khandelwal, Zand, Matosich, Veeraraghavan, Michelena, Li, Jagadeesh, Huang, Chawla, Huang, Chen, Garg, A, Silva, Bell, Zhang, Guo, Yu, Moshkovich, Wehrstedt, Khabsa, Avalani, Bhatt, Mankus, Hasson, Lennie, Reso, Groshev, Naumov, Lathi, Keneally, Liu, Seltzer, Valko, Restrepo, Patel, Vyatskov, Samvelyan, Clark, Macey, Wang, Hermoso, Metanat, Rastegari, Bansal, Santhanam, Parks, White, Bawa, Singhal, Egebo, Usunier, Mehta, Laptev, Dong, Cheng, Chernoguz, Hart, Salpekar, Kalinli, Kent, Parekh, Saab, Balaji, Rittner, Bontrager, Roux, Dollar, Zvyagina, Ratanchandani, Yuvraj, Liang, Alao, Rodriguez, Ayub, Murthy, Nayani, Mitra, Parthasarathy, Li, Hogan, Battey, Wang, Howes, Rinott, Mehta, Siby, Bondu, Datta, Chugh, Hunt, Dhillon, Sidorov, Pan, Mahajan, Verma, Yamamoto, Ramaswamy, Lindsay, Lindsay, Feng, Lin, Zha, Patil, Shankar, Zhang, Zhang, Wang, Agarwal, Sajuyigbe, Chintala, Max, Chen, Kehoe, Satterfield, Govindaprasad, Gupta, Deng, Cho, Virk, Subramanian, Choudhury,
  Goldman, Remez, Glaser, Best, Koehler, Robinson, Li, Zhang, Matthews, Chou, Shaked, Vontimitta, Ajayi, Montanez, Mohan, Kumar, Mangla, Ionescu, Poenaru, Mihailescu, Ivanov, Li, Wang, Jiang, Bouaziz, Constable, Tang, Wu, Wang, Wu, Gao, Kleinman, Chen, Hu, Jia, Qi, Li, Zhang, Zhang, Adi, Nam, Yu, Wang, Zhao, Hao, Qian, Li, He, Rait, DeVito, Rosnbrick, Wen, Yang, Zhao, and Ma]{llama}
Aaron Grattafiori, Abhimanyu Dubey, Abhinav Jauhri, Abhinav Pandey, Abhishek Kadian, Ahmad Al-Dahle, Aiesha Letman, Akhil Mathur, Alan Schelten, Alex Vaughan, Amy Yang, Angela Fan, Anirudh Goyal, Anthony Hartshorn, Aobo Yang, Archi Mitra, Archie Sravankumar, Artem Korenev, Arthur Hinsvark, Arun Rao, Aston Zhang, Aurelien Rodriguez, Austen Gregerson, Ava Spataru, Baptiste Roziere, Bethany Biron, Binh Tang, Bobbie Chern, Charlotte Caucheteux, Chaya Nayak, Chloe Bi, Chris Marra, Chris McConnell, Christian Keller, Christophe Touret, Chunyang Wu, Corinne Wong, Cristian~Canton Ferrer, Cyrus Nikolaidis, Damien Allonsius, Daniel Song, Danielle Pintz, Danny Livshits, Danny Wyatt, David Esiobu, Dhruv Choudhary, Dhruv Mahajan, Diego Garcia-Olano, Diego Perino, Dieuwke Hupkes, Egor Lakomkin, Ehab AlBadawy, Elina Lobanova, Emily Dinan, Eric~Michael Smith, Filip Radenovic, Francisco Guzmán, Frank Zhang, Gabriel Synnaeve, Gabrielle Lee, Georgia~Lewis Anderson, Govind Thattai, Graeme Nail, Gregoire Mialon, Guan Pang,
  Guillem Cucurell, Hailey Nguyen, Hannah Korevaar, Hu~Xu, Hugo Touvron, Iliyan Zarov, Imanol~Arrieta Ibarra, Isabel Kloumann, Ishan Misra, Ivan Evtimov, Jack Zhang, Jade Copet, Jaewon Lee, Jan Geffert, Jana Vranes, Jason Park, Jay Mahadeokar, Jeet Shah, Jelmer van~der Linde, Jennifer Billock, Jenny Hong, Jenya Lee, Jeremy Fu, Jianfeng Chi, Jianyu Huang, Jiawen Liu, Jie Wang, Jiecao Yu, Joanna Bitton, Joe Spisak, Jongsoo Park, Joseph Rocca, Joshua Johnstun, Joshua Saxe, Junteng Jia, Kalyan~Vasuden Alwala, Karthik Prasad, Kartikeya Upasani, Kate Plawiak, Ke~Li, Kenneth Heafield, Kevin Stone, Khalid El-Arini, Krithika Iyer, Kshitiz Malik, Kuenley Chiu, Kunal Bhalla, Kushal Lakhotia, Lauren Rantala-Yeary, Laurens van~der Maaten, Lawrence Chen, Liang Tan, Liz Jenkins, Louis Martin, Lovish Madaan, Lubo Malo, Lukas Blecher, Lukas Landzaat, Luke de~Oliveira, Madeline Muzzi, Mahesh Pasupuleti, Mannat Singh, Manohar Paluri, Marcin Kardas, Maria Tsimpoukelli, Mathew Oldham, Mathieu Rita, Maya Pavlova, Melanie Kambadur,
  Mike Lewis, Min Si, Mitesh~Kumar Singh, Mona Hassan, Naman Goyal, Narjes Torabi, Nikolay Bashlykov, Nikolay Bogoychev, Niladri Chatterji, Ning Zhang, Olivier Duchenne, Onur Çelebi, Patrick Alrassy, Pengchuan Zhang, Pengwei Li, Petar Vasic, Peter Weng, Prajjwal Bhargava, Pratik Dubal, Praveen Krishnan, Punit~Singh Koura, Puxin Xu, Qing He, Qingxiao Dong, Ragavan Srinivasan, Raj Ganapathy, Ramon Calderer, Ricardo~Silveira Cabral, Robert Stojnic, Roberta Raileanu, Rohan Maheswari, Rohit Girdhar, Rohit Patel, Romain Sauvestre, Ronnie Polidoro, Roshan Sumbaly, Ross Taylor, Ruan Silva, Rui Hou, Rui Wang, Saghar Hosseini, Sahana Chennabasappa, Sanjay Singh, Sean Bell, Seohyun~Sonia Kim, Sergey Edunov, Shaoliang Nie, Sharan Narang, Sharath Raparthy, Sheng Shen, Shengye Wan, Shruti Bhosale, Shun Zhang, Simon Vandenhende, Soumya Batra, Spencer Whitman, Sten Sootla, Stephane Collot, Suchin Gururangan, Sydney Borodinsky, Tamar Herman, Tara Fowler, Tarek Sheasha, Thomas Georgiou, Thomas Scialom, Tobias Speckbacher,
  Todor Mihaylov, Tong Xiao, Ujjwal Karn, Vedanuj Goswami, Vibhor Gupta, Vignesh Ramanathan, Viktor Kerkez, Vincent Gonguet, Virginie Do, Vish Vogeti, Vítor Albiero, Vladan Petrovic, Weiwei Chu, Wenhan Xiong, Wenyin Fu, Whitney Meers, Xavier Martinet, Xiaodong Wang, Xiaofang Wang, Xiaoqing~Ellen Tan, Xide Xia, Xinfeng Xie, Xuchao Jia, Xuewei Wang, Yaelle Goldschlag, Yashesh Gaur, Yasmine Babaei, Yi~Wen, Yiwen Song, Yuchen Zhang, Yue Li, Yuning Mao, Zacharie~Delpierre Coudert, Zheng Yan, Zhengxing Chen, Zoe Papakipos, Aaditya Singh, Aayushi Srivastava, Abha Jain, Adam Kelsey, Adam Shajnfeld, Adithya Gangidi, Adolfo Victoria, Ahuva Goldstand, Ajay Menon, Ajay Sharma, Alex Boesenberg, Alexei Baevski, Allie Feinstein, Amanda Kallet, Amit Sangani, Amos Teo, Anam Yunus, Andrei Lupu, Andres Alvarado, Andrew Caples, Andrew Gu, Andrew Ho, Andrew Poulton, Andrew Ryan, Ankit Ramchandani, Annie Dong, Annie Franco, Anuj Goyal, Aparajita Saraf, Arkabandhu Chowdhury, Ashley Gabriel, Ashwin Bharambe, Assaf Eisenman, Azadeh
  Yazdan, Beau James, Ben Maurer, Benjamin Leonhardi, Bernie Huang, Beth Loyd, Beto~De Paola, Bhargavi Paranjape, Bing Liu, Bo~Wu, Boyu Ni, Braden Hancock, Bram Wasti, Brandon Spence, Brani Stojkovic, Brian Gamido, Britt Montalvo, Carl Parker, Carly Burton, Catalina Mejia, Ce~Liu, Changhan Wang, Changkyu Kim, Chao Zhou, Chester Hu, Ching-Hsiang Chu, Chris Cai, Chris Tindal, Christoph Feichtenhofer, Cynthia Gao, Damon Civin, Dana Beaty, Daniel Kreymer, Daniel Li, David Adkins, David Xu, Davide Testuggine, Delia David, Devi Parikh, Diana Liskovich, Didem Foss, Dingkang Wang, Duc Le, Dustin Holland, Edward Dowling, Eissa Jamil, Elaine Montgomery, Eleonora Presani, Emily Hahn, Emily Wood, Eric-Tuan Le, Erik Brinkman, Esteban Arcaute, Evan Dunbar, Evan Smothers, Fei Sun, Felix Kreuk, Feng Tian, Filippos Kokkinos, Firat Ozgenel, Francesco Caggioni, Frank Kanayet, Frank Seide, Gabriela~Medina Florez, Gabriella Schwarz, Gada Badeer, Georgia Swee, Gil Halpern, Grant Herman, Grigory Sizov, Guangyi, Zhang, Guna
  Lakshminarayanan, Hakan Inan, Hamid Shojanazeri, Han Zou, Hannah Wang, Hanwen Zha, Haroun Habeeb, Harrison Rudolph, Helen Suk, Henry Aspegren, Hunter Goldman, Hongyuan Zhan, Ibrahim Damlaj, Igor Molybog, Igor Tufanov, Ilias Leontiadis, Irina-Elena Veliche, Itai Gat, Jake Weissman, James Geboski, James Kohli, Janice Lam, Japhet Asher, Jean-Baptiste Gaya, Jeff Marcus, Jeff Tang, Jennifer Chan, Jenny Zhen, Jeremy Reizenstein, Jeremy Teboul, Jessica Zhong, Jian Jin, Jingyi Yang, Joe Cummings, Jon Carvill, Jon Shepard, Jonathan McPhie, Jonathan Torres, Josh Ginsburg, Junjie Wang, Kai Wu, Kam~Hou U, Karan Saxena, Kartikay Khandelwal, Katayoun Zand, Kathy Matosich, Kaushik Veeraraghavan, Kelly Michelena, Keqian Li, Kiran Jagadeesh, Kun Huang, Kunal Chawla, Kyle Huang, Lailin Chen, Lakshya Garg, Lavender A, Leandro Silva, Lee Bell, Lei Zhang, Liangpeng Guo, Licheng Yu, Liron Moshkovich, Luca Wehrstedt, Madian Khabsa, Manav Avalani, Manish Bhatt, Martynas Mankus, Matan Hasson, Matthew Lennie, Matthias Reso, Maxim
  Groshev, Maxim Naumov, Maya Lathi, Meghan Keneally, Miao Liu, Michael~L. Seltzer, Michal Valko, Michelle Restrepo, Mihir Patel, Mik Vyatskov, Mikayel Samvelyan, Mike Clark, Mike Macey, Mike Wang, Miquel~Jubert Hermoso, Mo~Metanat, Mohammad Rastegari, Munish Bansal, Nandhini Santhanam, Natascha Parks, Natasha White, Navyata Bawa, Nayan Singhal, Nick Egebo, Nicolas Usunier, Nikhil Mehta, Nikolay~Pavlovich Laptev, Ning Dong, Norman Cheng, Oleg Chernoguz, Olivia Hart, Omkar Salpekar, Ozlem Kalinli, Parkin Kent, Parth Parekh, Paul Saab, Pavan Balaji, Pedro Rittner, Philip Bontrager, Pierre Roux, Piotr Dollar, Polina Zvyagina, Prashant Ratanchandani, Pritish Yuvraj, Qian Liang, Rachad Alao, Rachel Rodriguez, Rafi Ayub, Raghotham Murthy, Raghu Nayani, Rahul Mitra, Rangaprabhu Parthasarathy, Raymond Li, Rebekkah Hogan, Robin Battey, Rocky Wang, Russ Howes, Ruty Rinott, Sachin Mehta, Sachin Siby, Sai~Jayesh Bondu, Samyak Datta, Sara Chugh, Sara Hunt, Sargun Dhillon, Sasha Sidorov, Satadru Pan, Saurabh Mahajan,
  Saurabh Verma, Seiji Yamamoto, Sharadh Ramaswamy, Shaun Lindsay, Shaun Lindsay, Sheng Feng, Shenghao Lin, Shengxin~Cindy Zha, Shishir Patil, Shiva Shankar, Shuqiang Zhang, Shuqiang Zhang, Sinong Wang, Sneha Agarwal, Soji Sajuyigbe, Soumith Chintala, Stephanie Max, Stephen Chen, Steve Kehoe, Steve Satterfield, Sudarshan Govindaprasad, Sumit Gupta, Summer Deng, Sungmin Cho, Sunny Virk, Suraj Subramanian, Sy~Choudhury, Sydney Goldman, Tal Remez, Tamar Glaser, Tamara Best, Thilo Koehler, Thomas Robinson, Tianhe Li, Tianjun Zhang, Tim Matthews, Timothy Chou, Tzook Shaked, Varun Vontimitta, Victoria Ajayi, Victoria Montanez, Vijai Mohan, Vinay~Satish Kumar, Vishal Mangla, Vlad Ionescu, Vlad Poenaru, Vlad~Tiberiu Mihailescu, Vladimir Ivanov, Wei Li, Wenchen Wang, Wenwen Jiang, Wes Bouaziz, Will Constable, Xiaocheng Tang, Xiaojian Wu, Xiaolan Wang, Xilun Wu, Xinbo Gao, Yaniv Kleinman, Yanjun Chen, Ye~Hu, Ye~Jia, Ye~Qi, Yenda Li, Yilin Zhang, Ying Zhang, Yossi Adi, Youngjin Nam, Yu, Wang, Yu~Zhao, Yuchen Hao, Yundi
  Qian, Yunlu Li, Yuzi He, Zach Rait, Zachary DeVito, Zef Rosnbrick, Zhaoduo Wen, Zhenyu Yang, Zhiwei Zhao, and Zhiyu Ma.
\newblock The llama 3 herd of models, 2024.
\newblock URL \url{https://arxiv.org/abs/2407.21783}.

\bibitem[Hendrycks et~al.(2021)Hendrycks, Burns, Kadavath, Arora, Basart, Tang, Song, and Steinhardt]{mathhard}
Dan Hendrycks, Collin Burns, Saurav Kadavath, Akul Arora, Steven Basart, Eric Tang, Dawn Song, and Jacob Steinhardt.
\newblock Measuring mathematical problem solving with the math dataset.
\newblock \emph{arXiv preprint arXiv:2103.03874}, 2021.

\bibitem[Hu et~al.(2025)Hu, Liu, Xu, and Shen]{reinforce}
Jian Hu, Jason~Klein Liu, Haotian Xu, and Wei Shen.
\newblock Reinforce++: An efficient rlhf algorithm with robustness to both prompt and reward models, 2025.
\newblock URL \url{https://arxiv.org/abs/2501.03262}.

\bibitem[{Hugging Face}(2025)]{openr1}
{Hugging Face}.
\newblock Open r1: A fully open reproduction of deepseek-r1, January 2025.
\newblock URL \url{https://github.com/huggingface/open-r1}.

\bibitem[Jiang et~al.(2023)Jiang, Sablayrolles, Mensch, Bamford, Chaplot, de~las Casas, Bressand, Lengyel, Lample, Saulnier, Lavaud, Lachaux, Stock, Scao, Lavril, Wang, Lacroix, and Sayed]{mistral}
Albert~Q. Jiang, Alexandre Sablayrolles, Arthur Mensch, Chris Bamford, Devendra~Singh Chaplot, Diego de~las Casas, Florian Bressand, Gianna Lengyel, Guillaume Lample, Lucile Saulnier, Lélio~Renard Lavaud, Marie-Anne Lachaux, Pierre Stock, Teven~Le Scao, Thibaut Lavril, Thomas Wang, Timothée Lacroix, and William~El Sayed.
\newblock Mistral 7b, 2023.
\newblock URL \url{https://arxiv.org/abs/2310.06825}.

\bibitem[Jin et~al.(2020)Jin, Lin, and Li]{jin}
Zehao Jin, J.~Lin, and Siao-Fong Li.
\newblock Learning principle of least action with reinforcement learning.
\newblock \emph{ArXiv}, abs/2011.11891, 2020.
\newblock URL \url{https://api.semanticscholar.org/CorpusId:227151304}.

\bibitem[Kurita \& S{\o}gaard(2019)Kurita and S{\o}gaard]{kurita}
Shuhei Kurita and Anders S{\o}gaard.
\newblock Multi-task semantic dependency parsing with policy gradient for learning easy-first strategies.
\newblock In \emph{Annual Meeting of the Association for Computational Linguistics}, 2019.
\newblock URL \url{https://api.semanticscholar.org/CorpusId:174797942}.

\bibitem[Liu et~al.(2025{\natexlab{a}})Liu, Diao, Lu, Hu, Dong, Choi, Kautz, and Dong]{prorl}
Mingjie Liu, Shizhe Diao, Ximing Lu, Jian Hu, Xin Dong, Yejin Choi, Jan Kautz, and Yi~Dong.
\newblock Prorl: Prolonged reinforcement learning expands reasoning boundaries in large language models, 2025{\natexlab{a}}.
\newblock URL \url{https://arxiv.org/abs/2505.24864}.

\bibitem[Liu et~al.(2022)Liu, Flet-Berliac, and Brunskill]{liu}
Yao Liu, Yannis Flet-Berliac, and E.~Brunskill.
\newblock Offline policy optimization with eligible actions.
\newblock \emph{ArXiv}, abs/2207.00632, 2022.
\newblock URL \url{https://api.semanticscholar.org/CorpusId:250265118}.

\bibitem[Liu et~al.(2025{\natexlab{b}})Liu, Zhang, Zhu, Dong, Zhou, Shang, Yang, and Yang]{msoft}
Yifei Liu, Li~Lyna Zhang, Yi~Zhu, Bingcheng Dong, Xudong Zhou, Ning Shang, Fan Yang, and Mao Yang.
\newblock rstar-coder: Scaling competitive code reasoning with a large-scale verified dataset, 2025{\natexlab{b}}.
\newblock URL \url{https://arxiv.org/abs/2505.21297}.

\bibitem[Liu et~al.(2025{\natexlab{c}})Liu, Chen, Li, Qi, Pang, Du, Lee, and Lin]{drgrpo}
Zichen Liu, Changyu Chen, Wenjun Li, Penghui Qi, Tianyu Pang, Chao Du, Wee~Sun Lee, and Min Lin.
\newblock Understanding r1-zero-like training: A critical perspective, 2025{\natexlab{c}}.
\newblock URL \url{https://arxiv.org/abs/2503.20783}.

\bibitem[Mead et~al.(2025)Mead, Costen, Lacerda, and Hawes]{mead}
Harry Mead, Clarissa Costen, Bruno Lacerda, and Nick Hawes.
\newblock Return capping: Sample-efficient cvar policy gradient optimisation.
\newblock \emph{ArXiv}, abs/2504.20887, 2025.
\newblock URL \url{https://api.semanticscholar.org/CorpusId:278170796}.

\bibitem[Qwen et~al.(2025)Qwen, :, Yang, Yang, Zhang, Hui, Zheng, Yu, Li, Liu, Huang, Wei, Lin, Yang, Tu, Zhang, Yang, Yang, Zhou, Lin, Dang, Lu, Bao, Yang, Yu, Li, Xue, Zhang, Zhu, Men, Lin, Li, Tang, Xia, Ren, Ren, Fan, Su, Zhang, Wan, Liu, Cui, Zhang, and Qiu]{qwen}
Qwen, :, An~Yang, Baosong Yang, Beichen Zhang, Binyuan Hui, Bo~Zheng, Bowen Yu, Chengyuan Li, Dayiheng Liu, Fei Huang, Haoran Wei, Huan Lin, Jian Yang, Jianhong Tu, Jianwei Zhang, Jianxin Yang, Jiaxi Yang, Jingren Zhou, Junyang Lin, Kai Dang, Keming Lu, Keqin Bao, Kexin Yang, Le~Yu, Mei Li, Mingfeng Xue, Pei Zhang, Qin Zhu, Rui Men, Runji Lin, Tianhao Li, Tianyi Tang, Tingyu Xia, Xingzhang Ren, Xuancheng Ren, Yang Fan, Yang Su, Yichang Zhang, Yu~Wan, Yuqiong Liu, Zeyu Cui, Zhenru Zhang, and Zihan Qiu.
\newblock Qwen2.5 technical report, 2025.
\newblock URL \url{https://arxiv.org/abs/2412.15115}.

\bibitem[Shao et~al.(2024)Shao, Wang, Zhu, Xu, Song, Bi, Zhang, Zhang, Li, Wu, and Guo]{grpo}
Zhihong Shao, Peiyi Wang, Qihao Zhu, Runxin Xu, Junxiao Song, Xiao Bi, Haowei Zhang, Mingchuan Zhang, Y.~K. Li, Y.~Wu, and Daya Guo.
\newblock Deepseekmath: Pushing the limits of mathematical reasoning in open language models, 2024.
\newblock URL \url{https://arxiv.org/abs/2402.03300}.

\bibitem[Sohn et~al.(2021)Sohn, Lee, Choi, Seijen, Fatemi, and Lee]{sohn}
Sungryull Sohn, Sungtae Lee, Jongwook Choi, H.~V. Seijen, Mehdi Fatemi, and Honglak Lee.
\newblock Shortest-path constrained reinforcement learning for sparse reward tasks.
\newblock In \emph{International Conference on Machine Learning}, 2021.
\newblock URL \url{https://api.semanticscholar.org/CorpusId:235826326}.

\bibitem[Team et~al.(2025)Team, Kamath, Ferret, Pathak, Vieillard, Merhej, Perrin, Matejovicova, Ramé, Rivière, Rouillard, Mesnard, Cideron, bastien Grill, Ramos, Yvinec, Casbon, Pot, Penchev, Liu, Visin, Kenealy, Beyer, Zhai, Tsitsulin, Busa-Fekete, Feng, Sachdeva, Coleman, Gao, Mustafa, Barr, Parisotto, Tian, Eyal, Cherry, Peter, Sinopalnikov, Bhupatiraju, Agarwal, Kazemi, Malkin, Kumar, Vilar, Brusilovsky, Luo, Steiner, Friesen, Sharma, Sharma, Gilady, Goedeckemeyer, Saade, Feng, Kolesnikov, Bendebury, Abdagic, Vadi, György, Pinto, Das, Bapna, Miech, Yang, Paterson, Shenoy, Chakrabarti, Piot, Wu, Shahriari, Petrini, Chen, Lan, Choquette-Choo, Carey, Brick, Deutsch, Eisenbud, Cattle, Cheng, Paparas, Sreepathihalli, Reid, Tran, Zelle, Noland, Huizenga, Kharitonov, Liu, Amirkhanyan, Cameron, Hashemi, Klimczak-Plucińska, Singh, Mehta, Lehri, Hazimeh, Ballantyne, Szpektor, Nardini, Pouget-Abadie, Chan, Stanton, Wieting, Lai, Orbay, Fernandez, Newlan, yeong Ji, Singh, Black, Yu, Hui, Vodrahalli, Greff, Qiu,
  Valentine, Coelho, Ritter, Hoffman, Watson, Chaturvedi, Moynihan, Ma, Babar, Noy, Byrd, Roy, Momchev, Chauhan, Sachdeva, Bunyan, Botarda, Caron, Rubenstein, Culliton, Schmid, Sessa, Xu, Stanczyk, Tafti, Shivanna, Wu, Pan, Rokni, Willoughby, Vallu, Mullins, Jerome, Smoot, Girgin, Iqbal, Reddy, Sheth, Põder, Bhatnagar, Panyam, Eiger, Zhang, Liu, Yacovone, Liechty, Kalra, Evci, Misra, Roseberry, Feinberg, Kolesnikov, Han, Kwon, Chen, Chow, Zhu, Wei, Egyed, Cotruta, Giang, Kirk, Rao, Black, Babar, Lo, Moreira, Martins, Sanseviero, Gonzalez, Gleicher, Warkentin, Mirrokni, Senter, Collins, Barral, Ghahramani, Hadsell, Matias, Sculley, Petrov, Fiedel, Shazeer, Vinyals, Dean, Hassabis, Kavukcuoglu, Farabet, Buchatskaya, Alayrac, Anil, Dmitry, Lepikhin, Borgeaud, Bachem, Joulin, Andreev, Hardin, Dadashi, and Hussenot]{gemma}
Gemma Team, Aishwarya Kamath, Johan Ferret, Shreya Pathak, Nino Vieillard, Ramona Merhej, Sarah Perrin, Tatiana Matejovicova, Alexandre Ramé, Morgane Rivière, Louis Rouillard, Thomas Mesnard, Geoffrey Cideron, Jean bastien Grill, Sabela Ramos, Edouard Yvinec, Michelle Casbon, Etienne Pot, Ivo Penchev, Gaël Liu, Francesco Visin, Kathleen Kenealy, Lucas Beyer, Xiaohai Zhai, Anton Tsitsulin, Robert Busa-Fekete, Alex Feng, Noveen Sachdeva, Benjamin Coleman, Yi~Gao, Basil Mustafa, Iain Barr, Emilio Parisotto, David Tian, Matan Eyal, Colin Cherry, Jan-Thorsten Peter, Danila Sinopalnikov, Surya Bhupatiraju, Rishabh Agarwal, Mehran Kazemi, Dan Malkin, Ravin Kumar, David Vilar, Idan Brusilovsky, Jiaming Luo, Andreas Steiner, Abe Friesen, Abhanshu Sharma, Abheesht Sharma, Adi~Mayrav Gilady, Adrian Goedeckemeyer, Alaa Saade, Alex Feng, Alexander Kolesnikov, Alexei Bendebury, Alvin Abdagic, Amit Vadi, András György, André~Susano Pinto, Anil Das, Ankur Bapna, Antoine Miech, Antoine Yang, Antonia Paterson, Ashish
  Shenoy, Ayan Chakrabarti, Bilal Piot, Bo~Wu, Bobak Shahriari, Bryce Petrini, Charlie Chen, Charline~Le Lan, Christopher~A. Choquette-Choo, CJ~Carey, Cormac Brick, Daniel Deutsch, Danielle Eisenbud, Dee Cattle, Derek Cheng, Dimitris Paparas, Divyashree~Shivakumar Sreepathihalli, Doug Reid, Dustin Tran, Dustin Zelle, Eric Noland, Erwin Huizenga, Eugene Kharitonov, Frederick Liu, Gagik Amirkhanyan, Glenn Cameron, Hadi Hashemi, Hanna Klimczak-Plucińska, Harman Singh, Harsh Mehta, Harshal~Tushar Lehri, Hussein Hazimeh, Ian Ballantyne, Idan Szpektor, Ivan Nardini, Jean Pouget-Abadie, Jetha Chan, Joe Stanton, John Wieting, Jonathan Lai, Jordi Orbay, Joseph Fernandez, Josh Newlan, Ju~yeong Ji, Jyotinder Singh, Kat Black, Kathy Yu, Kevin Hui, Kiran Vodrahalli, Klaus Greff, Linhai Qiu, Marcella Valentine, Marina Coelho, Marvin Ritter, Matt Hoffman, Matthew Watson, Mayank Chaturvedi, Michael Moynihan, Min Ma, Nabila Babar, Natasha Noy, Nathan Byrd, Nick Roy, Nikola Momchev, Nilay Chauhan, Noveen Sachdeva, Oskar
  Bunyan, Pankil Botarda, Paul Caron, Paul~Kishan Rubenstein, Phil Culliton, Philipp Schmid, Pier~Giuseppe Sessa, Pingmei Xu, Piotr Stanczyk, Pouya Tafti, Rakesh Shivanna, Renjie Wu, Renke Pan, Reza Rokni, Rob Willoughby, Rohith Vallu, Ryan Mullins, Sammy Jerome, Sara Smoot, Sertan Girgin, Shariq Iqbal, Shashir Reddy, Shruti Sheth, Siim Põder, Sijal Bhatnagar, Sindhu~Raghuram Panyam, Sivan Eiger, Susan Zhang, Tianqi Liu, Trevor Yacovone, Tyler Liechty, Uday Kalra, Utku Evci, Vedant Misra, Vincent Roseberry, Vlad Feinberg, Vlad Kolesnikov, Woohyun Han, Woosuk Kwon, Xi~Chen, Yinlam Chow, Yuvein Zhu, Zichuan Wei, Zoltan Egyed, Victor Cotruta, Minh Giang, Phoebe Kirk, Anand Rao, Kat Black, Nabila Babar, Jessica Lo, Erica Moreira, Luiz~Gustavo Martins, Omar Sanseviero, Lucas Gonzalez, Zach Gleicher, Tris Warkentin, Vahab Mirrokni, Evan Senter, Eli Collins, Joelle Barral, Zoubin Ghahramani, Raia Hadsell, Yossi Matias, D.~Sculley, Slav Petrov, Noah Fiedel, Noam Shazeer, Oriol Vinyals, Jeff Dean, Demis Hassabis,
  Koray Kavukcuoglu, Clement Farabet, Elena Buchatskaya, Jean-Baptiste Alayrac, Rohan Anil, Dmitry, Lepikhin, Sebastian Borgeaud, Olivier Bachem, Armand Joulin, Alek Andreev, Cassidy Hardin, Robert Dadashi, and Léonard Hussenot.
\newblock Gemma 3 technical report, 2025.
\newblock URL \url{https://arxiv.org/abs/2503.19786}.

\bibitem[Xie et~al.(2024)Xie, Huang, Zhang, Yu, Chen, Lin, Li, Ghazi, and Kumar]{knk}
Chulin Xie, Yangsibo Huang, Chiyuan Zhang, Da~Yu, Xinyun Chen, Bill~Yuchen Lin, Bo~Li, Badih Ghazi, and Ravi Kumar.
\newblock On memorization of large language models in logical reasoning.
\newblock 2024.
\newblock URL \url{https://arxiv.org/abs/2410.23123}.

\bibitem[Xie et~al.(2025)Xie, Qiu, Gopinath, Lin, Sun, Wang, Potdar, and Dhingra]{apple}
Roy Xie, David Qiu, Deepak Gopinath, Dong Lin, Yanchao Sun, Chong Wang, Saloni Potdar, and Bhuwan Dhingra.
\newblock Interleaved reasoning for large language models via reinforcement learning, 2025.
\newblock URL \url{https://arxiv.org/abs/2505.19640}.

\bibitem[Yu et~al.(2025)Yu, Zhang, Zhu, Yuan, Zuo, Yue, Dai, Fan, Liu, Liu, Liu, Lin, Lin, Ma, Sheng, Tong, Zhang, Zhang, Zhang, Zhu, Zhu, Chen, Chen, Wang, Yu, Song, Wei, Zhou, Liu, Ma, Zhang, Yan, Qiao, Wu, and Wang]{dapo}
Qiying Yu, Zheng Zhang, Ruofei Zhu, Yufeng Yuan, Xiaochen Zuo, Yu~Yue, Weinan Dai, Tiantian Fan, Gaohong Liu, Lingjun Liu, Xin Liu, Haibin Lin, Zhiqi Lin, Bole Ma, Guangming Sheng, Yuxuan Tong, Chi Zhang, Mofan Zhang, Wang Zhang, Hang Zhu, Jinhua Zhu, Jiaze Chen, Jiangjie Chen, Chengyi Wang, Hongli Yu, Yuxuan Song, Xiangpeng Wei, Hao Zhou, Jingjing Liu, Wei-Ying Ma, Ya-Qin Zhang, Lin Yan, Mu~Qiao, Yonghui Wu, and Mingxuan Wang.
\newblock Dapo: An open-source llm reinforcement learning system at scale, 2025.
\newblock URL \url{https://arxiv.org/abs/2503.14476}.

\bibitem[Yu et~al.(2020)Yu, Jiang, Dong, and Feng]{reclor}
Weihao Yu, Zihang Jiang, Yanfei Dong, and Jiashi Feng.
\newblock Reclor: A reading comprehension dataset requiring logical reasoning, 2020.
\newblock URL \url{https://arxiv.org/abs/2002.04326}.

\bibitem[Yue et~al.(2025)Yue, Yuan, Yu, Zuo, Zhu, Xu, Chen, Wang, Fan, Du, Wei, Yu, Liu, Liu, Liu, Lin, Lin, Ma, Zhang, Zhang, Zhang, Zhu, Zhang, Liu, Wang, Wu, and Yan]{vapo}
Yu~Yue, Yufeng Yuan, Qiying Yu, Xiaochen Zuo, Ruofei Zhu, Wenyuan Xu, Jiaze Chen, Chengyi Wang, TianTian Fan, Zhengyin Du, Xiangpeng Wei, Xiangyu Yu, Gaohong Liu, Juncai Liu, Lingjun Liu, Haibin Lin, Zhiqi Lin, Bole Ma, Chi Zhang, Mofan Zhang, Wang Zhang, Hang Zhu, Ru~Zhang, Xin Liu, Mingxuan Wang, Yonghui Wu, and Lin Yan.
\newblock Vapo: Efficient and reliable reinforcement learning for advanced reasoning tasks, 2025.
\newblock URL \url{https://arxiv.org/abs/2504.05118}.

\end{thebibliography}
\bibliographystyle{iclr2026_conference}
}

\appendix
\pagebreak
\setcounter{page}{1}

\section{Appendix}
\subsection{Limitations and Future Work}
While our work establishes the Principle of Least Resistance as a fundamental law governing current models, it also illuminates several exciting frontiers for future investigation. Our experiments were conducted on models up to 12B parameters; an important open question is how this principle evolves at extreme scale, where the \enquote{difficulty} of our format hierarchies may become trivial, potentially altering the sequential learning dynamics. Furthermore, our investigation deliberately focused on reasoning domains with verifiable correctness to ensure empirical rigor. A valuable next step would be to characterize how this principle manifests in more subjective, open-ended tasks like creative writing or summarization, where the \enquote{simplest} path may be defined by stylistic or semantic factors beyond mere format complexity. Consequently, having identified and quantified this powerful bias, our work lays the theoretical groundwork for the crucial next phase of research: developing principled, targeted interventions. Future work could explore automated curriculum strategies that slowly increase complexity, or adaptive reward-shaping methods that penalize cognitive shortcuts, with the ultimate goal of architecting optimization landscapes where the path of true, effortful cognition is also, fundamentally, the path of least resistance.

\subsection{Experimental Setup}
\subsubsection{Optimization Hyperparameters}

\begin{table}[h]
\centering
\caption{Core hyperparameters for the GRPO, DAPO, and REINFORCE++ training runs.}
\label{tab:hyperparams}
\begin{tabular}{lccc}
\toprule
\textbf{Parameter} & \textbf{GRPO} & \textbf{DAPO} & \textbf{REINFORCE++} \\
\midrule
\multicolumn{4}{l}{\textit{Optimizer Settings}} \\
Learning Rate (`lr`) & 5e-6 & 5e-6 & 5e-5 \\
Adam $\beta_1$ / $\beta_2$ & 0.9 / 0.99 & 0.9 / 0.99 & 0.9 / 0.99 \\
Weight Decay & 0.1 & 0.1 & 0.1 \\
LR Scheduler & cosine & cosine & cosine \\
Warmup Ratio & 0.1 & 0.1 & 0.1 \\
Max Grad Norm & 0.15 & 0.15 & 0.15 \\
\midrule
\multicolumn{4}{l}{\textit{Training \& Batching}} \\
Effective Batch Size & 8 & 8 & 8 \\
Epochs & 3 & 3 & 3 \\
Precision & bfloat16 & bfloat16 & bfloat16 \\
\midrule
\multicolumn{4}{l}{\textit{PO-Specific Settings}} \\
Loss Type & \textit{dr\_grpo} & \textit{dr\_grpo} & - \\
$\epsilon$ / $\epsilon_{high}$ & 0.2 / 0.2 & 0.2 / 0.28 & - \\
KL Beta ($\beta$) & 0.1 / 0.3 & 0.1 / 0.3 & - \\
Generations per Prompt & 5 & 5 & 1 \\
PPO Epochs & - & - & 2 \\
Advantage Normalization & - & - & True \\
Reward Normalization & - & - & True \\
\midrule
\multicolumn{4}{l}{\textit{LoRA Settings}} \\
Rank (``r'') & 64 & 64 & 64 \\
Alpha ($\alpha$) & 128 & 128 & 128 \\
Target Modules & All Linear Proj. & All Linear Proj. & All Linear Proj. \\
\bottomrule
\end{tabular}
\end{table}

In large-scale PO, preventing policy collapse is paramount. We therefore adopted a small, conservative \textbf{learning rate of $5e^{-6}$ for our value-based methods (GRPO, DAPO)}, a standard choice that ensures smooth and consistent policy updates. For the actor-only \textbf{REINFORCE++}, we used a slightly higher $5e^{-5}$, which is common for this class of algorithms to ensure a sufficiently strong learning signal. Across all optimizers, we used a \textbf{cosine annealing scheduler with a $0.1$ warmup ratio and a strict max grad norm of $0.15$}. This combination is a powerful recipe for stability: the warmup allows the optimizer to find its footing, the cosine decay prevents divergence late in training, and the gradient clipping acts as a crucial safeguard against catastrophic updates.

For our primary algorithm, \textbf{GRPO}, we specifically chose the ``Done Right'' ($dr\_grpo$) variant. This was a deliberate decision to ensure the integrity of our findings. Standard GRPO can be susceptible to biases from varying response lengths and question difficulties. The $dr\_grpo$ loss incorporates corrections for these factors, providing a cleaner and more reliable learning signal. This ensures that the "Principle of Least Resistance" we observe is a true feature of the optimization landscape, not an artifact of a noisy or biased algorithm.

Our implementation of \textbf{DAPO} builds directly on this stable GRPO foundation, incorporating its key innovation: decoupled clipping. By setting \textbf{$\epsilon_{high}=0.28$}, we provide a larger ceiling for the importance ratio, granting the optimizer more freedom to increase the probability of low-probability (exploratory) tokens. This technique is designed to combat entropy collapse and was a crucial component in enabling the long, stable training runs required for our analysis.

All of the training runs were done in a parameter efficient fashion with LoRA. By targeting all linear projection layers with a moderate rank ($\mathbf{r}=64$) and a scaled alpha ($\mathbf{\alpha}=128$), we created an efficient and expressive adaptation method. This allowed us to effectively steer these large models without the prohibitive cost of full fine-tuning. To be memory efficient with the REINFORCE++ training, which requires a separate reference model, we leveraged different LoRA adapter heads with the same base model to create the policy model and the reference model.

\subsubsection{Computational Resources}

All experiments reported in this paper were conducted on a private cluster of NVIDIA GPUs. Specifically, our setup consisted of $3\text{x}$ \textbf{NVIDIA RTX A6000 GPUs}, each with $48\text{GB}$ of VRAM, and a single \textbf{NVIDIA H100 GPU} with $80\text{GB}$ of VRAM. The use of $\text{bf16}$ precision and standard optimization techniques such as \textbf{gradient checkpointing} was essential for fitting our larger model training runs within this memory footprint.
A typical experimental run for a single model on a single dataset required approximately \textbf{7.5 GPU hours} to complete. The full suite of experiments, including all models, algorithms, and the adversarial ablations amounted to a total of approximately \textbf{200 GPU hours}. We report these figures to provide a transparent account of the computational resources required to replicate our findings and to contextualize the scale of our empirical investigation.

\subsubsection{Reward Functions}
At the core of our investigation is the design of the reward functions that shape the optimization landscape. All our experiments are built upon a combination of a primary \textit{Correctness Reward} and several \textit{Format Rewards}. We define these formally below. Let $y$ be a model generation for a given prompt, and let $a$ be the ground-truth answer.

\paragraph{Correctness Reward ($R_{\text{correct}}$)}
The primary objective for any reasoning task is to arrive at the correct final answer. We define a correctness reward that is contingent on the model's output being both parsable and numerically equivalent to the ground-truth. We first define an extraction function, $\text{ExtractBoxed}(y)$, which retrieves the content within the final \verb|\boxed{}| environment in the generation $y$. The correctness reward is then given by:

\begin{equation}
    R_{\text{correct}}(y, a) = 
    \begin{cases} 
        +3.0 & \text{if } \text{Verify}(\text{ExtractBoxed}(y), a) \text{ is True} \\
        -5.0 & \text{if } \text{Verify}(\text{ExtractBoxed}(y), a) \text{ is False} \\
        -5.0 & \text{if } \text{ExtractBoxed}(y) \text{ is not parsable}
    \end{cases}
\end{equation}

where the function $\text{Verify}(\cdot, \cdot)$ checks for mathematical equivalence between the extracted response and the ground-truth answer. The large penalty for incorrectness provides a strong signal to avoid hallucination and guessing.

\paragraph{Composite Format Reward ($R_{\text{composite}}$)}
For our initial ``Liberating Experiment'', we designed a flexible format reward that accepts a wide range of reasoning structures. This function provides a positive reward if the generation $y$ matches the regular expression for either an interleaved ``think-then-answer'' structure or a direct `answer`-only format. Let $\mathcal{F}_{\text{composite}}$ be the set of all strings matching this permissive format. The reward is defined as:

\begin{equation}
    R_{\text{composite}}(y) = 
    \begin{cases} 
        +3.0 & \text{if } y \in \mathcal{F}_{\text{composite}} \\
        -4.0 & \text{otherwise}
    \end{cases}
\end{equation}

The total reward for this experiment is a linear combination of the correctness and composite format rewards.

\paragraph{Hierarchical Format Rewards ($R_{\text{hierarchy}}$)}
To formalize the "Principle of Least Resistance," we designed a set of three nested, Matryoshka-style format requirements, $f_1, f_2, f_3$, representing ascending levels of structural complexity.

\begin{itemize}
    \item \textbf{Easy Format ($f_1$):} The innermost requirement, satisfied if the generation simply contains a boxed answer. Let $\mathcal{F}_1$ be the set of strings matching \verb|^.*\\boxed\{.+\}.*$|.
    \begin{equation}
        r(y, f_1) = \begin{cases} +1.0 & \text{if } y \in \mathcal{F}_1 \\ -4.0 & \text{otherwise} \end{cases}
    \end{equation}

    \item \textbf{Medium Format ($f_2$):} A superset of $f_1$, requiring the boxed answer to be enclosed within `\verb|<answer>|` tags. Let $\mathcal{F}_2$ be the set of strings matching \verb|^.*<answer>.*\\boxed\{.+\}.*</answer>.*$|.
    \begin{equation}
        r(y, f_2) = \begin{cases} +2.0 & \text{if } y \in \mathcal{F}_2 \\ -4.0 & \text{otherwise} \end{cases}
    \end{equation}

    \item \textbf{Hard Format ($f_3$):} The most encompassing format, requiring the full `think-then-answer` structure. Let $\mathcal{F}_3$ be the set of strings matching \verb|^<think>.*</think><answer>.*\\boxed\{.+\}.*</answer>$|.
    \begin{equation}
        r(y, f_3) = \begin{cases} +4.0 & \text{if } y \in \mathcal{F}_3 \\ -4.0 & \text{otherwise} \end{cases}
    \end{equation}
\end{itemize}

For our ``Reward Hierarchy'' experiment, the reward magnitudes for $r(y, f_1)$, $r(y, f_2)$, and $r(y, f_3)$ were all set to a constant value, $R_{\text{max}}$, to isolate the effect of complexity. For the ``Geometric Incentive Gambit,'' we used the scaled values of 1.0, 2.0, and 4.0 as defined above to create a strong incentive gradient. The total reward for these experiments, $R_{\text{hierarchy}}(y)$, is the sum of the rewards for all satisfied format conditions, ensuring that a generation matching $f_3$ also receives the rewards for $f_2$ and $f_1$. For the ``Mutually Exclusive Choice'' experiment, the total reward is simply the reward for the single, highest-complexity format that is exclusively matched.

\paragraph{Conflicting Format Rewards ($R_{\text{conflict}}$)}
For our final adversarial test, the "Mutually Exclusive Choice" experiment, it was crucial to design format requirements that were truly disjoint, ensuring that a single generation \textbf{$y$} could not satisfy more than one condition. This eliminates the possibility of shared sub-problems and forces the optimizer into a clean cost-benefit analysis. We achieved this through careful construction of the regular expressions for each format. The reward function, $R_{\text{conflict}}(y)$, provides a constant positive reward $R_\text{max}$ if and only if s\textbf{$y$} exclusively matches one of the following formats, \textcolor{blue}{$f_1$}, \textcolor{ForestGreen}{$f_2$}, or \textcolor{orange}{$f_3$}.

\begin{itemize}
    \item \textbf{Easy Format ($f_1$):} Defined by the regex \verb|^.*\\boxed\{.+\}$|. This format is satisfied if the generation ends *immediately* after the closing brace of a $\text{\\boxed\{\}}$ environment. The `\$' anchor is critical; it asserts that no characters, including the `\verb|</answer>|` tag, can follow the boxed answer.

    \item \textbf{Medium Format ($f_2$):} Defined by the regex - 

        \verb|^(?:<answer>.+?\\boxed\{.+\}<\/answer>)$|  
    . This format requires the entire generation to be a single, valid \verb|<answer>| block that contains a boxed answer somewhere within it. The \textbf{\^} and \textbf{\$} anchors ensure that the generation must begin with `\verb|<answer>|` and end with \verb|</answer>|, with no preceding \verb|<think>| tags.

    \item \textbf{Hard Format ($f_3$):} Defined by the regex - 

        \verb|^<think>.+<\/think><answer>.+\\boxed\{.+\}<\/answer>$|    
    . This format mandates the strict \textit{think-then-answer} structure. The \^ anchor ensures the generation must start with a \verb|<think>| tag.
\end{itemize}

\textbf{Justification for Mutual Exclusivity: }
The disjoint nature of these formats is guaranteed by the careful use of start-of-string \texttt{\^} and end-of-string `\$' anchors in the regular expressions. Let us walk through why a single generation \textbf{$y$} cannot satisfy more than one condition:

\ding{182} A generation matching \textcolor{orange}{$f_3$} cannot match \textcolor{ForestGreen}{$f_2$} because it must start with \verb|<think>|, whereas \textcolor{ForestGreen}{$f_2$} must start with \verb|<answer>|.

\ding{183} A generation matching \textcolor{ForestGreen}{$f_2$} cannot match \textcolor{blue}{$f_1$} because it must end with \verb|</answer>|, whereas \textcolor{blue}{$f_1$} must end immediately after the \} of the boxed expression.

\ding{184} Similarly, a generation matching \textcolor{orange}{$f_3$} cannot match \textcolor{blue}{$f_1$} because it must also end with \verb|</think>|, violating the end-of-string requirement of \textcolor{blue}{$f_1$}.

This design ensures that the three reward paths are truly independent, forcing the optimizer to commit to a single strategy. The resulting convergence on textcolor{blue}{$f_1$}, as shown in the main paper, is therefore an unconfounded demonstration of its intrinsic preference for the most easily learnable format, even when other, equally rewarded paths are available.

\subsection{System Prompts}
The system prompt is a critical component of the experimental setup, as it frames the task for the model and can subtly influence its behavior. For our investigation, it was paramount to design prompts that were as neutral as possible, defining the valid formats without biasing the model towards any particular one. Below, we detail the prompts used for our core experiments and the rationale behind their design.

\paragraph{Prompt for Composite and Hierarchical Reward Experiments}
For all experiments involving a choice between multiple valid formats (Section \ref{sec:law_seq_opt}, Section \ref{sec:quantify_resistance}), we used the following system prompt. This prompt was engineered to explicitly grant the model freedom and define the available tools (\texttt{<think>}, \texttt{<answer>}, \verb|\boxed{}|), while deliberately avoiding any language that would suggest one format is preferable to another.

\begin{tcolorbox}[
    title={System Prompt:},
    fonttitle=\bfseries,
    coltitle=gray!15,
    sharp corners,
    boxrule=1pt,
    rounded corners,
    arc=5mm, 
    colback=gray!10, 
    borderline west={2pt}{0pt}{black} 
]
You are a helpful assistant, and you will be providing factually correct
responses to the user queries. While responding, if at any point you feel
like you are not comfortable answering the question, you can choose to
think in an internal dialogue and enclose your thoughts between \texttt{<think>}\texttt{</think>}
tags, and then when you are confident again, you can resume answering by
enclosing your answer between \texttt{<answer>}\texttt{</answer>} tags. You can choose to
think whenever you feel appropriate, and then answer when you are confident.
Note that this also means that you can think a bit, answer that portion, and
start thinking again when you need to, which means you can have interleaving
\texttt{<think>}...\texttt{</think>} and \texttt{<answer>}...\texttt{</answer>} tags in your response. Enclose your
final response to the query in \verb|\boxed{}|.
\end{tcolorbox}

\paragraph{Design Rationale: } The key to this prompt is its permissive and non-prescriptive language. Phrases like \textit{you can choose to think}, \textit{whenever you feel appropriate} and the explicit mention of an \enquote{interleaving} structure are designed to signal to the model that all valid formats defined by our composite and hierarchical rewards are acceptable. The prompt provides no incentive, explicit or implicit, to prefer a minimal response. The emergence of the ``Cognitive Shortcut'' under this neutral prompt is therefore a stronger signal of the optimizer's intrinsic bias.

\paragraph{Prompt for Strict think-then-answer Baseline}
For our baseline experiments and the ``Mutually Exclusive Choice'' experiment (where \textcolor{orange}{$f_3$} was the hard format), we used a standard, more restrictive prompt. 

\begin{tcolorbox}[
    title={System Prompt:},
    fonttitle=\bfseries,
    coltitle=gray!15,
    sharp corners,
    boxrule=1pt,
    rounded corners,
    arc=5mm, 
    colback=gray!10, 
    borderline west={2pt}{0pt}{black} 
]
A conversation between User and Assistant. The user asks a question, and
the Assistant solves it. The assistant first needs to think between
\texttt{<think>}...\texttt{</think>}, and then respond between \texttt{<answer>}...\texttt{</answer>} tags. The final response within the answer must be enclosed within \verb|\boxed{}|.\end{tcolorbox}

\paragraph{Design Rationale: } This prompt is explicitly prescriptive. The phrase \textit{first needs to think... and then respond} removes all ambiguity and enforces a single, sequential computational path. This serves as our control condition, representing the standard methodology against which we measure the behavior of the liberated policy. By using such a clear and unambiguous instruction, we ensure that any differences in learned behavior can be directly attributed to the change in the reward function and the freedom it grants, rather than to any ambiguity in the prompt itself.

\subsection{Related Work}
\label{sec:related-work}

A significant body of research has explored ``shortcut learning'' \cite{sohn, kurita, jin} where models find simple, unintended solutions. However, this work has largely focused on undesirable behavior as an artifact of the problem setup; such as sparse data, inefficient algorithms, or task-specific heuristics rather than as a fundamental preference of the optimizer itself. Our work pivots from this external view to an internal one, identifying and formalizing the \emph{Principle of Least Resistance} as an  intrinsic bias in policy optimization, a critical blind spot in the context of aligning increasingly autonomous models.

\paragraph{Shortcuts as Inefficiency or Statistical Artifacts: }
A common perspective treats shortcuts as obstacles to finding a truly optimal policy. For instance, in sparse-reward RL, \cite{sohn} propose a ``shortest-path constraint'' to prevent an agent from taking physically redundant actions, like moving back and forth. Their ``shortest path'' is the geometrically optimal trajectory, representing the desired solution. This stands in stark contrast to our findings, where the ``Path of Least Resistance'' is the cognitively simplest strategy, representing the problem of a degenerate, suboptimal policy. Similarly, in the offline RL setting, \cite{liu} identify ``propensity overfitting,'' a statistical shortcut where the policy games the importance sampling estimator by avoiding initial states with poor data coverage. Their work masterfully diagnoses a shortcut born from the interaction between a policy and its evaluator under finite data. Our work, however, reveals a shortcut that persists even in the online, on-policy setting with full data access, demonstrating that the bias is not statistical but is inherent to the \emph{optimizer's search dynamics}. These works aim to fix the external conditions to help a neutral optimizer; we show the optimizer is not neutral.

\paragraph{Shortcuts as Algorithmic Flaws or Features: }
Another line of work addresses the mechanics of specific algorithms or tasks. \cite{mead} tackle the ``blindness to success'' in risk-averse CVaR optimization, where the algorithm is forced by its mathematical definition to discard high-performing trajectories. Their solution, Return Capping, enables the optimizer to learn from all experiences. The failure they address is the algorithm being unable to see success; the failure we identify is the optimizer choosing to ignore a more successful, complex strategy in favor of an easier one. This distinction is crucial: one is an algorithmic flaw, the other a behavioral bias.

\cite{jin} show that an RL agent can learn to follow the ``Principle of Least Action''---a fundamental law of physics describing optimal trajectories---by engineering a corresponding reward function. Here again, the path of ``least action'' is the ground-truth, optimal solution. Our work provides a compelling counterpoint: the ``Principle of Least Resistance'' describes the optimizer's preference not for the most efficient solution, but for the most learnable one, which is often a shortcut.

In summary, prior work has studied shortcuts as circumstantial problems to be engineered away. Our contribution is the identification of a universal, predictable principle governing the optimizer's behavior. In the current research landscape, where models are granted unprecedented freedom to explore complex reward landscapes, understanding this intrinsic bias is no longer a secondary concern---it is the central challenge for building robust and reliable reasoning systems. Our work shifts the focus from fixing the \emph{conditions} of optimization to understanding the \emph{optimizer itself}.

\begin{figure*}[h!]
    \centering
    \begin{myuser}{User}
A ship traveling along a river has covered $24 \mathrm{~km}$ upstream and $28 \mathrm{~km}$ downstream. For this journey, it took half an hour less than for traveling $30 \mathrm{~km}$ upstream and $21 \mathrm{~km}$ downstream, or half an hour more than for traveling $15 \mathrm{~km}$ upstream and $42 \mathrm{~km}$ downstream, assuming that both the ship and the river move uniformly.

Determine the speed of the ship in still water and the speed of the river.
    \end{myuser}
    \begin{mybox}{Qwen-7B + OpenR1-Math220K}
        \begin{mybox}{$r_\text{strict}$}
<think>Let’s think step by step...</think>
<answer>... \boxed{4, 10}</answer>
        \end{mybox}
        \begin{mybox}{$r_\text{composite}$}
<answer>Let's denote the speed of the ship in still water as $s$ and the speed of the river as … \$10 km/h and the speed of the river is 2 km/h. \boxed{10, 2}</answer>
        \end{mybox}
    \end{mybox}
\caption{}
    \vspace{-5mm}
    \label{fig:intro_demo}
\end{figure*}

\begin{figure*}[h!]
    \centering
    \begin{myuser}{User}
    Let $AB$ be a chord of the unit circle $\odot O$. If the area of $\odot O$ is exactly equal to the area of the square with side $AB$, then $\angle AOB=$ $\qquad$ (to 0.001 degree).
    \end{myuser}
    \begin{mybox}{Qwen-7B + OpenR1-Math220K}
        \begin{mybox}{$r_\text{strict}$}
  <think>Okay, so the problem is about a chord AB on the unit circle ... So, setting these equal ... </think>
  <answer>$124.806^{\circ}$</answer>
        \end{mybox}
        \begin{mybox}{$r_\text{composite}$}
        <answer>First, we need to find the area of the unit circle ... Therefore, the angle AOB is approximately 65.427 degrees, rounded to 0.001 degree. \boxed{65.427}</answer>
        \end{mybox}
    \end{mybox}
\caption{}
    \vspace{-5mm}
    \label{fig:intro_demo}
\end{figure*}

\begin{figure*}[h!]
    \centering
    \begin{myuser}{User}
Four boys bought a boat for $\$60$. The first boy paid one half of the sum of the amounts paid by the other boys; the second boy paid one third of the sum of the amounts paid by the other boys; and the third boy paid one fourth of the sum of the amounts paid by the other boys. How much did the fourth boy pay? 

\textbf{Answer: } $13$
    \end{myuser}
    \begin{mybox}{Qwen-7B + OpenR1-Math220K}
        \begin{mybox}{$r_\text{strict}$}
<think>Let $x_1, x_2, x_3, x_4$ be the amounts paid by the four boys. We are given that $x_1 + x_2 + x_3 + x_4 = 60$.
The first boy paid one half of the sum of the amounts paid by the other boys, so $x_1 = \frac{1}{2}(x_2 + x_3 + x_4)$.
The second boy paid one third of the sum of the amounts paid by the other boys, so $x_2 = \frac{1}{3}(x_1 + x_3 + x_4)$... </think>
<answer>The amounts paid by the four boys are $20, 15, 12, 13$.
Let's check the conditions:
$x_1 = 20$, $x_2 = 15$, $x_3 = 12$, $x_4 = 13$.
$x_1 = \frac{1}{2}(x_2 + x_3 + x_4) = \frac{1}{2}(15 + 12 + 13) = \frac{1}{2}(40) = 20$.
$x_2 = \frac{1}{3}(x_1 + x_3 + x_4) = \frac{1}{3}(20 + 12 + 13) = \frac{1}{3}(45) = 15$.
$x_3 = \frac{1}{4}(x_1 + x_2 + x_4) = \frac{1}{4}(20 + 15 + 13) = \frac{1}{4}(48) = 12$.
The conditions are satisfied.

Final Answer: The final answer is $\boxed{13}$</answer>
        \end{mybox}
        \begin{mybox}{$r_\text{composite}$}
<answer>\boxed{10}</answer>
        \end{mybox}
    \end{mybox}
\caption{}
    \vspace{-5mm}
    \label{fig:intro_demo}
\end{figure*}

\end{document}